# THE THEORETICAL AND PRACTICAL ASPECTS OF FUZZY REASONING

(1)

## An Analysis of General Fuzzy Logic and Fuzzy Reasoning Method

Kwak Son Il

Professor, Ph.D. College of Computer Science, **Kim Il Sung** University,

Pyongyang, DPR of Korea, Kwak Son Il <ryongnam18@yahoo.com>

Hyon Gyong Il,

Professor, Ph.D. College of Computer Science, **Kim Il Sung** University,

Pyongyang, DPR of Korea

Pak Ji Min

Academician, Professor, Ph.D. Department of Automation Engineering,
Kim Chaek University of Technology, Pyongyang, DPR of Korea

In this article, we describe the fuzzy logic, fuzzy language and algorithms as the basis of fuzzy reasoning, one of the intelligent information processing method, and then describe the general fuzzy reasoning method.

## 1.1. Fuzzy Logic

### 1.1.1. Concept of fuzzy logic

Fuzzy logic has its real significance in several aspect such as fuzzy language, fuzzy control theory, information processing, computer science, medical diagnosis and so on.

This fuzzy logic has been developed since making the classical binary logic vague. For a long time, mathematical logic has been existed as a pure theory, but in the late of 1930's the practical value appeared obviously. First, it was used in design of switch circuit and then became the important basis of computer science and automatic control theory as computer developed. However, because a preposition in binary logic has only 2 values: true and false, it is inconvenient for emulation of the human thinking.

Objective things are not absolutes only with true and false, and there is much transitional state between true and false and this shows vagueness. Because human being have intelligence processing the objects with vagueness smoothly, so they can think generally and parallel and then they have ability applying the summarization, abstraction, intuition and thinking creatively. Therefore the task to search the formulation problem of natural language was suggested and under this premise fuzzy logic also has been developed.

Fuzzy logic is the branch that generalized the use of connection operation in mathematical logic and selection of value in truth table, and its development shows its powerful vitality by combined with the computer science. In computer manufacturing, mainly 《minimization of logic operation》 is studied and in program, mainly study on approximation reasoning is used in artificial intelligence.

The characteristics of fuzzy logic are as follows:

① Truth value is vague, that is a proposition may have any value between range [0, 1].

② Truth value can be generated by given grammar.

③ Using truth value can have meaning according to the grammatical rule.

④ It has meaning that the truth value can be changed as local logic.

⑤ Reasoning rule is natural, not certain.

Here, ② and ③ are the problems related with fuzzy language, and ①, ④ and ⑤ are the one related with fuzzy logic, so we have to understand the meaning considering the contents described in fuzzy logic.

### 1.1.2. Fuzzy Proposition

In binary logic, each proposition has only two values: true 《1》 and false 《0》. A proposition is expressed with the symbol and formulas having the standing rules, so study on formulation of logic proposition has became to convert to the change of certain symbols and formulas.

Fuzzy proposition is a kind of symbolic logic and can be said the generalization of general proposition. In fuzzy logic, the value which a fuzzy proposition has doesn't express the 《true》 and 《false》, but expresses the degree of them. Therefore, the truth value can be all the real number ranging in interval [0, 1]. And fuzzy logic is widely used in controlling the complicated production process, using the switch theory, information process, approximation reasoning, and fuzzy language.

[Definition 1.1.1] Fuzzy proposition is the sentence having the vagueness. In other words, proposition is defined as the proposition which truth value is in [0, 1].

In general, proposition is expressed in the form of 《x is A》, where x is called subject, and A is called predicate. The difference between the fuzzy proposition and general proposition is that the its predicate is expressed in the form of fuzzy set.

Fuzzy proposition is denoted as symbols $\underset{\sim}{P}, \underset{\sim}{Q}, \underset{\sim}{R}, \cdots$.

(Example 1.1.1)

$\underset{\sim}{P}$ : 《x is a small positive number》

$\underset{\sim}{Q}$ : 《Miss. Pak is very strong.》

$\underset{\sim}{R}$ : 《It is very hot today》

In the above example, 《is a small positive number》, 《is very strong》 and 《is very hot》 are all the fuzzy concepts and they are expressed as fuzzy set. Therefore fuzzy proposition is corresponded to the fuzzy set.

Arithmetic of truth value of fuzzy proposition is the one of the membership function.

Fuzzy union $\vee$, fuzzy intersection $\wedge$ and complement $\neg$ are defined as (1.1.1), (1.1.2) and (1.1.3).

$$\underset{\sim}{P} \vee \underset{\sim}{Q} = \max(\underset{\sim}{P}, \underset{\sim}{Q}) \qquad (1.1.1)$$

$$\underset{\sim}{P} \wedge \underset{\sim}{Q} = \min(\underset{\sim}{P},\ \underset{\sim}{Q}) \tag{1.1.2}$$

$$\neg \underset{\sim}{P} = 1 - \underset{\sim}{P}. \tag{1.1.3}$$

Then we can obtain the following result for the logic operations $\vee$, $\wedge$, $\neg$.

Here, we refer $\underset{\sim}{P}$, $\underset{\sim}{Q}$, $\underset{\sim}{R}$ to the fuzzy proposition having value in [0, 1], and denote $\wedge$ to $\cdot$, $\neg$ to $-$.

① Commutativity

$$\underset{\sim}{P} \vee \underset{\sim}{Q} = \underset{\sim}{Q} \vee \underset{\sim}{P},\ \underset{\sim}{P} \cdot \underset{\sim}{Q} = \underset{\sim}{Q} \cdot \underset{\sim}{P} \tag{1.1.4}$$

② Idempotency

$$\underset{\sim}{P} \vee \underset{\sim}{P} = \underset{\sim}{P},\ \underset{\sim}{P} \cdot \underset{\sim}{P} = \underset{\sim}{P} \tag{1.1.5}$$

③ Involution

$$\overline{\overline{\underset{\sim}{P}}} = \underset{\sim}{P} \tag{1.1.6}$$

④ Associativity

$$\underset{\sim}{P} \vee (\underset{\sim}{Q} \vee \underset{\sim}{R}) = (\underset{\sim}{P} \vee \underset{\sim}{Q}) \vee \underset{\sim}{R} \tag{1.1.7}$$

$$\underset{\sim}{P} \cdot (\underset{\sim}{Q} \cdot \underset{\sim}{R}) = (\underset{\sim}{P} \cdot \underset{\sim}{Q}) \cdot \underset{\sim}{R} \tag{1.1.8}$$

⑤ Absorption

$$\underset{\sim}{P} \vee (\underset{\sim}{P} \cdot \underset{\sim}{Q}) = \underset{\sim}{P},\quad \underset{\sim}{P} \cdot (\underset{\sim}{P} \vee \underset{\sim}{Q}) = \underset{\sim}{P} \tag{1.1.9}$$

⑥ Distributivity

$$\underset{\sim}{P} \cdot (\underset{\sim}{Q} \vee \underset{\sim}{R}) = \underset{\sim}{P} \cdot \underset{\sim}{Q} \vee \underset{\sim}{P} \cdot \underset{\sim}{R} \tag{1.1.10}$$

$$\underset{\sim}{P} \vee (\underset{\sim}{Q} \cdot \underset{\sim}{R}) = (\underset{\sim}{P} \vee \underset{\sim}{Q}) \cdot (\underset{\sim}{P} \vee \underset{\sim}{R}). \tag{1.1.11}$$

⑦ De Morgan's law

$$\overline{(\underset{\sim}{P} \vee \underset{\sim}{Q})} = \overline{\underset{\sim}{P}} \cdot \overline{\underset{\sim}{Q}}, \quad \overline{(\underset{\sim}{P} \cdot \underset{\sim}{Q})} = \overline{\underset{\sim}{P}} \vee \overline{\underset{\sim}{Q}} \tag{1.1.12}$$

⑧ Zero element

$$0 \vee \underset{\sim}{P} = \underset{\sim}{P}, \quad 0 \cdot \underset{\sim}{P} = 0 \tag{1.1.13}$$

⑨ Identity

$$1 \vee \underset{\sim}{P} = 1, \quad 1 \cdot \underset{\sim}{P} = \underset{\sim}{P} \tag{1.1.14}$$

⑩ Complementary law

$$\overline{\underset{\sim}{P}} \vee \underset{\sim}{P} \neq 1, \quad \overline{\underset{\sim}{P}} \cdot \underset{\sim}{P} \neq 0 \tag{1.1.15}$$

In some cases, we may denote the complementary law as

$$\overline{\underset{\sim}{P}} \vee \underset{\sim}{P} \geq \frac{1}{2}, \quad \overline{\underset{\sim}{P}} \cdot \underset{\sim}{P} \leq \frac{1}{2} \tag{1.1.16}$$

From the above equations, the following equations are held.

$$\overline{\underset{\sim}{P}} \cdot \underset{\sim}{P} = \overline{\underset{\sim}{P}} \cdot \underset{\sim}{P} \cdot (\overline{\underset{\sim}{Q}} \vee \underset{\sim}{Q}) = \overline{\underset{\sim}{P}} \cdot \underset{\sim}{P} \cdot \overline{\underset{\sim}{Q}} \vee \overline{\underset{\sim}{P}} \cdot \underset{\sim}{P} \cdot \underset{\sim}{Q} \tag{1.1.17}$$

$$\overline{\underset{\sim}{P}} \vee \underset{\sim}{P} = \overline{\underset{\sim}{P}} \vee \underset{\sim}{P} \vee (\overline{\underset{\sim}{Q}} \cdot \underset{\sim}{Q}) = (\overline{\underset{\sim}{P}} \vee \underset{\sim}{P} \vee \overline{\underset{\sim}{Q}}) \cdot (\overline{\underset{\sim}{P}} \vee \underset{\sim}{P} \vee \underset{\sim}{Q}) \tag{1.1.18}$$

If the truth value of fuzzy proposition $\underset{\sim}{P}$ on each point of object set $X$ is not less than $\lambda$, that is

$$\underset{\sim}{P} \geq \lambda \tag{1.1.19}$$

then fuzzy proposition $\underset{\sim}{P}$ is defined as $\lambda-$always true proposition where $\lambda \in [0, 1]$.

Especially when $\lambda = 1$, it becomes the always true proposition.

### 1.1.3. Basic knowledge of fuzzy reasoning.

In binary logic, reasoning sentence 《If ~ then ~》 is the conditional proposition and is denoted as symbol 《$P \to Q$》.

The truth value is as follows.

| $P$ | 1 | 1 | 0 | 0 |
|---|---|---|---|---|
| $Q$ | 1 | 0 | 1 | 0 |
| $P \to Q$ | 1 | 0 | 1 | 1 |

Because the truth value table of proposition 《$P \to Q$》 and the 《$\neg P \vee Q$》 are the same, (1.1.20) is held.

$$P \to Q = \neg P \vee Q \qquad (1.1.20)$$

When $\underset{\sim}{P}$ and $\underset{\sim}{Q}$ are fuzzy proposition, $\underset{\sim}{P} \to \underset{\sim}{Q}$ is called approximation reasoning sentence (fuzzy reasoning sentence).

For example, if

$\underset{\sim}{P}$: 《It is clear.》

$\underset{\sim}{Q}$: 《It is warm. 》

, then 《$\underset{\sim}{P} \sim \underset{\sim}{Q}$》 means 《If it is clear, it is warm》, and 《$\underset{\sim}{Q} \sim \underset{\sim}{P}$》 means 《If it is warm, it is clear.》

$$P \to Q = \neg P \vee Q$$

Generalizing the $P \to Q = \neg P \vee Q$ as the fuzzy proposition,

$$\underset{\sim}{P} \to \underset{\sim}{Q} = \neg \underset{\sim}{P} \vee \underset{\sim}{Q}$$

Therefore, when proposition 《$x$ change the litmus paper into red color》 is expressed using the set by using a syllogistic; we should consider the objective space to two factors: one is 《chemical property》 and the other is 《color》.

Now, consider the approximation reasoning (fuzzy reasoning) related with two object space $X,\ Y$

$$\underset{\sim}{P} \to \underset{\sim}{Q} \tag{1.1.21}$$

To simplify, first consider the general proposition $P \to Q$. We can consider relation $P \to Q$ on $X \times Y$. The characteristic function is

$$(P \to Q)(x,\ y) = (P(x) \wedge Q(y)) \vee (1 - P(x)) \tag{1.1.22}$$

where $x \in X,\ y \in Y$, and $P(x)$ and $Q(y)$ have only 0 or 1.

Generalizing this equation in the form of fuzzy reasoning $\underset{\sim}{P} \to \underset{\sim}{Q}$ is the fuzzy relation on $X \times Y$ and its membership function is expressed as (1.1.22).

$$(\underset{\sim}{P} \to \underset{\sim}{Q})(x,\ y) = (\underset{\sim}{P}(x) \wedge \underset{\sim}{Q}(y)) \vee (1 - \underset{\sim}{P}(x)) \tag{1.1.22}$$

where

$$\underset{\sim}{P}(x),\ \underset{\sim}{Q}(y) \in [0,\ 1].$$

(Example 1.1.2) Two fuzzy propositions

《$\underset{\sim}{P}$ : $x$ is small.》 and 《$\underset{\sim}{Q}$: $y$ is large.》

$\underset{\sim}{P} \to \underset{\sim}{Q}$ means 《If $x$ is small, then $y$ is large》.

Refer the object space to $X = Y = \{1,\ 2,\ 3,\ 4,\ 5\}$, and 《small》 and 《large》 are the fuzzy sets on two object space. The result of calculation by fuzzy reasoning formula is as follows.

| $\underset{\sim}{P} \to \underset{\sim}{Q}$ | $y$: 1 | 2 | 3 | 4 | 5 |
|---|---|---|---|---|---|
| $x$: 1 | 0 | 0 | 0 | 0.5 | 1 |
| 2 | 0.5 | 0.5 | 0.5 | 0.5 | 0.5 |
| 3 | 1 | 1 | 1 | 1 | 1 |
| 4 | 1 | 1 | 1 | 1 | 1 |
| 5 | 1 | 1 | 1 | 1 | 1 |

This determine a fuzzy relation matrix $\underset{\sim}{R}$.

That is,

$$\underset{\sim}{R} = \begin{pmatrix} 0 & 0 & 0 & 0.5 & 1 \\ 0.5 & 0.5 & 0.5 & 0.5 & 0.5 \\ 1 & 1 & 1 & 1 & 1 \\ 1 & 1 & 1 & 1 & 1 \\ 1 & 1 & 1 & 1 & 1 \end{pmatrix}$$

This fuzzy relation $\underset{\sim}{R}$ means 《If $x$ is small, then $y$ is large》.

When we know the fuzzy relation 《If $x$ is small, then $y$ is large.》, then we can do the following fuzzy reasoning.

《If $x$ is small then $y$ is large.》

《If $x$ is more or less small then how is $y$?》

where Matrix of fuzzy relation $\underset{\sim}{P}$ 《If $x$ is small then $y$ is large.》 And object space $X$ and $Y$ are the same above.

《Is more or less small》 $(x) = 1/1 + 0.4/2 + 0.2/3$,

and denote it in the form of fuzzy vector.

$$\underset{\sim}{A} = (1,\ 0.4,\ 0.2,\ 0,\ 0)$$

$$\underset{\sim}{B} = \underset{\sim}{A} \circ \underset{\sim}{R} = (1\ \ 0.4\ \ 0.2\ \ 0\ \ 0) \circ \begin{pmatrix} 0 & 0 & 0 & 0.5 & 1 \\ 0.5 & 0.5 & 0.5 & 0.5 & 0.5 \\ 1 & 1 & 1 & 1 & 1 \\ 1 & 1 & 1 & 1 & 1 \\ 1 & 1 & 1 & 1 & 1 \end{pmatrix} =$$

$$= (0.4\ \ 0.4\ \ 0.4\ \ 0.5\ \ 1)$$

In the aspect of fuzzy conversion, the output of 《how is $y$?》 is as follows.

$$\underset{\sim}{B} = 0.4/1 + 0.4/2 + 0.4/3 + 0.5/4 + 1/5$$

This result shows 《$y$ is more or less large》 and that this is agreed with the human thinking method.

### 1.1.4 Fuzzy logical expression

Fuzzy logical expression is the generalization of general logic expression. The arithmetic rules of logic expression or propositional expression consist of Boolean algebra and the one of fuzzy logical expression is

$$\left.\begin{aligned} \underset{\sim}{F} \vee (\neg \underset{\sim}{F}) \neq 1 \\ \underset{\sim}{F} \wedge (\neg \underset{\sim}{F}) \neq 0 \end{aligned}\right\} \qquad (1.1.23)$$

For $\underset{\sim}{F}$ and the others satisfy the Boolean algebra.

To simplify the symbolic expression, we denote symbol 《$\wedge$》 to 《$\cdot$》, 《$\vee$》 to 《$+$》 and 《$\neg \underset{\sim}{F}$》 to 《$\overline{\underset{\sim}{F}}$》.

[Definition 1.1.2] Refer $x_1,\ x_2,\ \cdots,\ x_n$ to the fuzzy values [0, 1], then

$\underset{\sim}{F}(x_1, x_2, \cdots, x_n)$: fuzzy logical expression is the mapping

$$\underset{\sim}{F}:\ \underbrace{[0,\ 1] \times [0,\ 1] \times \cdots \times [0,\ 1]}_{n} \to [0,\ 1] \qquad (1.1.24)$$

Refer fuzzy logical expression $\underset{\sim}{F}$: $(x_1,\ x_2,\ \cdots,\ x_n)$ to $f-$formula.

The $f-$formula satisfies the following properties.

① Number 0 and 1 are $f-$formulas.

② Fuzzy variable $x_i$ is $f-$formula itself.

③ If $\underset{\sim}{F_1}, \underset{\sim}{F_2}$ are $f-$formulas, $\underset{\sim}{F_1} + \underset{\sim}{F_2}, \underset{\sim}{F_1} \cdot \underset{\sim}{F_2}$ are also $f-$formulas.

④ If $\underset{\sim}{F}$ is $f-$formula, then $\overline{\underset{\sim}{F}}$ is also $f-$formula.

⑤ Formula obtained by applying the above properties ①−④ for several times is also $f-$formula.

We refer the set of fuzzy logical expressions to F, and define the truth value function $T$ as the mapping $T$: F $\rightarrow [0,\ 1]$

Then, if $\underset{\sim}{F} \in$ F, $T(\underset{\sim}{F})$ expresses the true value of $\underset{\sim}{F}$.

The following properties are held.

① $T(\overline{\underset{\sim}{F}}) = 1 - T(\underset{\sim}{F})$

② $T(\underset{\sim}{F_1} + \underset{\sim}{F_2}) = \max(T(\underset{\sim}{F_1}),\ T(\underset{\sim}{F_2})) = T(\underset{\sim}{F_1}) + T(\underset{\sim}{F_2})$

③ $T(\underset{\sim}{F_1} + \underset{\sim}{F_2}) = \min(T(\underset{\sim}{F_1}),\ T(\underset{\sim}{F_2})) = T(\underset{\sim}{F_1}) \cdot T(\underset{\sim}{F_2})$

④ $T(\underset{\sim}{F_1} \rightarrow \underset{\sim}{F_2}) = \min(1,\ 1 - T(\underset{\sim}{F_1}) + T(\underset{\sim}{F_2}))$

In the above, we considered about the $\lambda-$always true proposition. When $\lambda = \frac{1}{2}$, the following is defined.

[Definition 1.1.3] For $f-$formula $\underset{\sim}{F}$, although the variables of $\underset{\sim}{F}$ are given any values, if $T(\underset{\sim}{F}) \geq \frac{1}{2}$ then $\underset{\sim}{F}$ is called fuzzy always true, and If $T(\underset{\sim}{F}) < \frac{1}{2}$ then $\underset{\sim}{F}$ is called fuzzy contradiction.

(Example 1.1.3) Considering the above definition for $\underset{\sim}{F} = \underset{\sim}{x} + \overline{\underset{\sim}{x}}$,

$$T(\underset{\sim}{F}) = \max(T(\underset{\sim}{x}), T(\overline{\underset{\sim}{x}})) = \begin{cases} T(\underset{\sim}{x}), & T(\underset{\sim}{x}) \geq \frac{1}{2} \\ 1 - T(\underset{\sim}{x}), & T(\underset{\sim}{x}) < \frac{1}{2} \end{cases} \tag{1.1.25}$$

Therefore, although $T(\underset{\sim}{x})$ has any value, always $T(\underset{\sim}{F}) \geq \frac{1}{2}$. So $\underset{\sim}{F}$ is fuzzy always true.

According to (1.1.25) for $\underset{\sim}{F} = \underset{\sim}{x} \cdot \overline{\underset{\sim}{x}}$

$$T(\underset{\sim}{F}) = \min(T(\underset{\sim}{x}),\ T(\overline{\underset{\sim}{x}})) = \begin{cases} T(\underset{\sim}{x}), & T(\underset{\sim}{x}) < \frac{1}{2} \\ 1 - T(\underset{\sim}{x}), & T(\underset{\sim}{x}) \geq \frac{1}{2} \end{cases}$$

Therefore, always $T(\underset{\sim}{F}) < \frac{1}{2}$, and $\underset{\sim}{F}$ is fuzzy contradiction.

To consider the properties of $f-$formula, we introduce the concept of clause and phrase as two conduct fuzzy logical expression.

We call a fuzzy variable $\underset{\sim}{x_i}$ and its negative character.

We denote a character $\underset{\sim}{L_i}$, then the logical add of characters $\underset{\sim}{L_1}, \underset{\sim}{L_2}, \cdots, \underset{\sim}{L_p}$

$$\underset{\sim}{L_1} + \underset{\sim}{L_2} + \cdots + \underset{\sim}{L_p} \tag{1.1.26}$$

is called clause and logical expression

$$\underset{\sim}{L_1} \cdot \underset{\sim}{L_2} \cdot \cdots \cdot \underset{\sim}{L_n} \tag{1.1.27}$$

is called phrase.

When the logical expression $\underset{\sim}{F}$ is expressed like

$$\underset{\sim}{F} = \underset{\sim}{\phi_1} + \underset{\sim}{\phi_2} + \cdots + \underset{\sim}{\phi_p},\ P \geq 1 \tag{1.1.28}$$

, we call $\underset{\sim}{F}$ fuzzy addition normal form.

When the following equation is held, given phrase $\underset{\sim}{C_i}$ , $\underset{\sim}{F}$ is called fuzzy multiplication normal form.

$$\underset{\sim}{F} = \underset{\sim}{C_1} \cdot \underset{\sim}{C_2} \cdot \cdots \cdot \underset{\sim}{C_p},\ P \geq 1 \tag{1.1.29}$$

In the binary logic, when logical expression *F* is held for any *T* as follows, it is called always true and contradiction respectively..

$$T(F) = 1 \tag{1.1.30}$$

$$T(F) = 0 \tag{1.1.31}$$

Then, consider which relation exists between the always true and contradiction in binary logic and fuzzy always true and fuzzy contradiction.

① It is necessary and sufficient that clause $\underset{\sim}{C}$ contains complement pair $(\underset{\sim}{x_i}, \underset{\sim}{\bar{x}_i})$ for clause $\underset{\sim}{C}$ is fuzzy always true.

② It is necessary and sufficient that clause $\underset{\sim}{\phi}$ contains complement pair $(\underset{\sim}{x_i}, \underset{\sim}{\bar{x}_i})$ for clause $\underset{\sim}{\phi}$ is fuzzy contradiction.

③ It is necessary and sufficient that every clause $\underset{\sim}{\phi_i}$ is fuzzy contradiction for fuzzy addition normal form

$$\underset{\sim}{F} = \underset{\sim}{\phi_1} + \underset{\sim}{\phi_2} + \cdots + \underset{\sim}{\phi_p} \qquad (1.1.32)$$

is fuzzy contradiction.

④ It is necessary and sufficient that every clauses $\underset{\sim}{C_i}$ is fuzzy contradiction for fuzzy multiplication normal form

$$\underset{\sim}{F} = \underset{\sim}{C_1} \cdot \underset{\sim}{C_2} \cdot \cdots \cdot \underset{\sim}{C_p} \qquad (1.1.33)$$

is fuzzy always true.

⑤ Condition that is necessary and sufficient for fuzzy logical expression $\underset{\sim}{F}$ is fuzzy always true is that the logical expression is always true in binary logic, and necessary and sufficient condition for $\underset{\sim}{F}$ is fuzzy contradiction is that it is contradiction in binary logic.

# 1.2 Analysis of Fuzzy Logic Function

## 1.2.1 Concepts of fuzzy logic function.

Denoted closed interval [0, 1] as *V,* logical intersection and logical sum in fuzzy logic can be considered as the two variable-functions from $V \times V$ to *V,* and logical negative one variable function from *V* to *V*.

In general, function from $V^n$ to *V*

$$F\colon V^n \to V \qquad (1.2.1)$$

is called n variable fuzzy function.

Those expressed as logic expression among the fuzzy function are called fuzzy logic expression.

Here logic expression is the equation consisted of combination of variable $x_i (i=\overline{1,\ n})$ and logic symbols $\vee, \cdot, \neg$, and variable $x_i$ is the fuzzy variable representing fuzzy proposition $\underset{\sim}{x}_i$.

(Example 1.2.1) $F(x_1,\ x_2,\ x_3) = (x_1 \vee \bar{x}_2 \cdot x_3) \cdot x_2 \vee x_1 \cdot \bar{x}_2 \cdot \bar{x}_3$

For any element $x = (x_1, x_2, x_3)$ in $V^3$, when $x = (0.1,\ \ 0.8,\ \ 0.4)$ is held, the following equation is held.

$$F(x) = (0.1 \vee (1-0.8) \cdot 0.4) \cdot 0.8 \vee 0.1 \cdot (1-0.8) \cdot (1-0.4) =$$

$$= (0.1 \vee 0.2 \cdot 0.4) \cdot 0.8 \vee 0.1 \cdot 0.2 \cdot 0.6 = (0.1 \vee 0.2) \cdot 0.8 \vee 0.1 \cdot 0.2 \cdot 0.6 =$$

$$= 0.2 \cdot 0.8 \vee 0.1 \cdot 0.2 \cdot 0.6 = 0.2 \vee 0.1 \cdot 0.2 \cdot 0.6 = 0.2 \cdot 0.2 \cdot 0.6 = 0.2$$

As you can see in the example, logic expression and logic function are the same in essence.

To simplify the consideration, we suppose that the true value of $F(a)$ $T(F(a))$ is the value of $F(a)$ itself, given $a \in V^n$.

Logical intersection of several characters that don't contain certain character and its rest elements at the same time is called single term, and logical intersection of several characters that contain certain variable and its rest elements at the same time is called complement term.

Single term and complement term are called term. And complement term with all the variables is called complement minimum term.

For example, term $x_1 \cdot x_2 \cdot \bar{x}_2 \cdot x_3$ on $V^3$ is complement minimum term, but $x_1 \cdot x_2 \cdot \bar{x}_2$ is not complement minimum term.

Given two fuzzy logic expressions $F_1$ and $F_2$, when the following (1.2.2) is held for any element *a,* we can describe that $F_1$ contains $F_2$, and equation (1.2.3) is held.

$$F_1(a) \geq F_2(a) \qquad (1.2.2)$$

$$F_1 \geq F_2 \qquad (1.2.3)$$

Especially, in (1.2.2), when $\alpha$ and $\beta$ are terms, Iff every character in $\alpha$ is in $\beta$, $\alpha$ contains $\beta$. Therefore, when $\alpha$ contains $\beta$, $\alpha + \beta = \alpha$ and $\beta$ can be omitted.

For example,

$$x_1 \cdot x_2 \cdot x_3 + x_1 \cdot x_2 = x_1 \cdot x_2 \cdot x_3$$

For any fuzzy logic function *F*, the formula obtained by expanding all complement terms as the sum of complement minimum terms is called fuzzy main addition normal form of *F*.

For example, fuzzy logic function on $V^3$

$$F = \overline{x}_1 \cdot x_2 + \overline{x}_1 \cdot x_2 \cdot \overline{x}_2 + x_1 \cdot x_2 \cdot x_3$$

, $\overline{x}_1 \cdot x_2 \cdot \overline{x}_2$ are complement terms. This can be expanded as the sum of the complement minimum terms as

$$\overline{x}_1 \cdot x_2 \cdot \overline{x}_2 = \overline{x}_1 \cdot x_2 \cdot \overline{x}_2 \cdot (x_3 + \overline{x}_3) = \overline{x}_1 \cdot x_2 \cdot \overline{x}_2 \cdot x_3 + \overline{x}_1 \cdot x_2 \cdot \overline{x}_2 \cdot \overline{x}_3$$

Therefore,

$$F = \overline{x}_1 \cdot x_2 + \overline{x}_1 \cdot x_2 \cdot \overline{x}_2 \cdot x_3 + \overline{x}_1 \cdot x_2 \cdot \overline{x}_2 \cdot \overline{x}_3 + x_1 \cdot x_2 \cdot x_3 = \overline{x}_1 \cdot x_2 + x_1 \cdot x_2 \cdot x_3$$

This is fuzzy main addition normal form of *F*.

Refer *F* to fuzzy logic function, and $\alpha$ to term, and when the (1.2.4) is held for each element $a$ of $V^n$, term $\alpha$ is called fuzzy closure of $F$. Especially, when $\alpha$ omitted any character is not fuzzy closure of $F$, $\alpha$ is called fuzzy item. This is denoted as FPI.

$$\alpha(a) \le F(a) \tag{1.2.4}$$

The form with the least characters among the addition normal form of fuzzy logic functions is called fuzzy simplest form.

Here, the following property is held.

When $F$ is the fuzzy main addition normal form, all single items in $F$ are the FPI of $F$.

By using the above property, given the fuzzy addition normal form $F$, we can find FPI of single item $F$. That is, if single item $\alpha$ in $F$ isn't contained in other single item, it has already been FPI of $F$. So, it is the problem to find the FPI of complement item. For this, we introduce the fuzzy consistency.

Refer $\alpha, \beta$ to item respectively, fuzzy consistency set of $\alpha$ and $\beta$ $FC(\alpha, \beta)$ is defined as follows.

Given a certain variable $x_i$, when $\alpha = x_i \cdot \alpha_0 (\bar{x}_i \not\geq \alpha_0),\ \beta = \bar{x}_i \cdot \beta_0 (x_i \not\geq \beta_0)$ or $\alpha = \bar{x}_i \cdot \alpha_0 (x_i \not\geq \alpha_0),\ \beta = x_i \cdot \beta_0 (\bar{x}_i \not\geq \beta_0)$

1° If $\alpha_0 \cdot \beta_0$ is complement item, then $\alpha_0 \beta_0 \in FC(\alpha, \beta)$

2° If $\alpha_0 \cdot \beta_0$ is simple term, then $\alpha_0 \cdot \beta_0 \cdot x_j \cdot \bar{x}_i \in FC(\alpha, \beta), (j \neq i)$

3° $FC(\alpha, \beta)$ consists of only complement items of 1° and 2°.

And the same character is replicated as fuzzy consistency, replicated characters should be omitted.

For instance, refer $\alpha,\ \beta$ to the items on $V^3$.

① If $\alpha = x_1 \cdot \bar{x}_2,\ \beta = \bar{x}_1 \cdot x_2$, then $FC(\alpha,\ \beta) = \{x_1 \cdot \bar{x}_1,\ x_2 \cdot \bar{x}_2\}$

② If $\alpha = x_1 \cdot x_2,\ \beta = x_1 \cdot \bar{x}_2$, then $FC(\alpha,\ \beta) = \{x_1 \cdot \bar{x}_1,\ x_1 \cdot x_3 \cdot \bar{x}_3\}$

③ If $\alpha = x_1,\ \beta = x_1$, then $FC(\alpha,\ \beta) = \{x_2 \cdot \bar{x}_2,\ x_3 \cdot \bar{x}_3\}$

Then, the following properties are held.

If fuzzy addition normal form $F = \alpha_1 + \alpha_2 + \cdots + \alpha_n$ consists of sum of all FPI of *F*,

1° Item contained in the other item doesn't exist.

2° Either fuzzy consistency of any two items doesn't exist, or it is contained in other item $\alpha_i (i = \overline{1,\ n})$.

From the above property, refer *F* to any fuzzy logic function, we can obtain computing algorithm obtaining all FPI of *F*.

Computing algorithm

1° Expand *F* as the fuzzy addition normal form..

2° Omit all items contained other item.

3° Obtain the fuzzy consistency for any two items. In case whether fuzzy consistency doesn't exist or is contained in other item, add the obtained fuzzy consistency and go to 2°.

4° The obtained formula is the sum of all FPI.

(Example 1.2.2) Calculate FPI of fuzzy logic function on $V^4$

$$F = x_1 \cdot \bar{x}_1 \cdot x_2 \cdot \bar{x}_3 + x_1 \cdot \bar{x}_1 \cdot x_2 \cdot \bar{x}_4 + x_1 \cdot \bar{x}_1 \cdot x_2 \cdot x_3 \cdot x_4 .$$

First, for step 1°, *F* is the fuzzy addition normal form. For step 2°, an item contained in the other term doesn't exist.

Next, refer

$$\alpha_1 = x_1 \cdot \bar{x}_1 \cdot x_2 \cdot \bar{x}_3$$
$$\alpha_2 = x_1 \cdot \bar{x}_1 \cdot x_2 \cdot \bar{x}_4$$
$$\alpha_3 = x_1 \cdot \bar{x}_1 \cdot x_2 \cdot x_3 \cdot x_4$$

and obtain the fuzzy consistency according to the step 3°, then

$$FC(\alpha_1,\ \alpha_2) = \phi$$
$$FC(\alpha_1,\ \alpha_3) = \{x_1 \cdot \bar{x}_1 \cdot x_2 \cdot x_4\}$$
$$FC(\alpha_1,\ \alpha_3) = \{x_1 \cdot \bar{x}_1 \cdot x_2 \cdot x_3\}$$

Because these fuzzy consistencies are not contained in other item, the following equation is held, adding them to *F*.

$$F = x_1 \cdot \bar{x}_1 \cdot x_2 \cdot \bar{x}_3 + x_1 \cdot \bar{x}_1 \cdot x_2 \cdot \bar{x}_4 + x_1 \cdot \bar{x}_1 \cdot x_2 \cdot x_3 \cdot x_4 + \\ + x_1 \cdot \bar{x}_1 \cdot x_2 \cdot x_4 + x_1 \cdot \bar{x}_1 \cdot x_2 \cdot x_3$$

Repeating the same step for *F*, $x_1 \cdot \bar{x}_1 \cdot x_2$ are obtained as FPI.

As we can see on the above consideration, one fuzzy logic function is expressed as FPI to decrease the number of characters. That is, fuzzy simplest form is expressed as the sum of FPI.

There are two cases in FPI of fuzzy logic function *F*, one is in simplest form of *F*, and the other is not in any simplest form. Some properties related with this are described as follows.

If FPI $\alpha$ is in all simplest form of fuzzy logic function *F*, $\alpha$ is called indispensable item and the other is called dispensable item.

From this definition, the following 2 properties are held.

1° If a certain single item $\alpha$ is FPI of fuzzy logic function *F*, $\alpha$ is indispensable item of *F*.

2° If $\gamma$ is fuzzy consistency of $\alpha$ and $\beta$ when single items $\alpha$, $\beta$ and complement item $\gamma$ are FPI of fuzzy logic function F, then $\gamma$ is dispensable item of F.

From the above properties, we can realize that it is not necessary to obtain the fuzzy consistency of single items to obtain the simplest form of fuzzy logic function.

Computing algorithm

1° Obtain all FPIs obtained by calculating the fuzzy consistency of items which at least one is complement item, and refer them to $\alpha_1, \alpha_2, \cdots, \alpha_n$.

2° Calculate the fuzzy main addition normal form. Refer the single item to $\beta_1, \beta_2, \cdots, \beta_m$, and find the minimum combination containing the complement minimum items $\gamma_1, \gamma_2, \cdots, \gamma_l$ in $\alpha_1, \alpha_2, \cdots, \alpha_n$, and then refer it $\alpha_1', \alpha_2', \cdots, \alpha_s'$.

3° $F = \beta_1 + \beta_2 + \cdots + \beta_m + \alpha_1' + \alpha_2' + \cdots + \alpha_s'$ is the simplest form of F.

(Example 1.2.3) Obtain the simplest form of

$F = \bar{x}_2 \cdot \bar{x}_4 + x_1 \cdot x_2 \cdot \bar{x}_3 + \bar{x}_1 \cdot x_2 \cdot x_4 + x_1 \cdot \bar{x}_2 \cdot x_3 \cdot x_4 + x_1 \cdot \bar{x}_1 \cdot \bar{x}_2 \cdot x_3 \cdot x_4$.

By step 1°, obtained FPI of complement item,

$$\alpha_1 = x_1 \cdot \bar{x}_1 \cdot \bar{x}_2$$
$$\alpha_2 = x_1 \cdot \bar{x}_1 \cdot \bar{x}_3 .$$
$$\alpha_3 = x_1 \cdot \bar{x}_1 \cdot x_4$$

In step 2°, simple item is

$$\beta_1 = \bar{x}_2 \cdot \bar{x}_4$$
$$\beta_2 = x_1 \cdot x_2 \cdot \bar{x}_3$$
$$\beta_3 = x_1 \cdot \bar{x}_2 \cdot x_3 \cdot x_4$$

and complement minimum item is

$$\gamma_1 = x_1 \cdot \bar{x}_1 \cdot \bar{x}_2 \cdot \bar{x}_3 \cdot x_4$$

, because $F$ has already been fuzzy addition normal form.

By step $3°$, minimum combination of $\alpha_i (i = 1, 2, 3)$ containing $\gamma_1$ is a certain $\alpha_i$. By step $4°$, simplest form of $F$ is

$$F = \bar{x}_2 \cdot \bar{x}_4 + x_1 \cdot x_2 \cdot \bar{x}_3 + \bar{x}_1 \cdot x_2 \cdot x_4 + x_1 \cdot \bar{x}_2 \cdot x_3 \cdot x_4 + \begin{Bmatrix} x_1 \cdot \bar{x}_1 \cdot \bar{x}_2 \\ x_1 \cdot \bar{x}_1 \cdot \bar{x}_3 \\ x_1 \cdot \bar{x}_1 \cdot x_4 \end{Bmatrix}.$$

As you can see, the simplest form may be not one.

Next, consider the monotonousness of fuzzy logic function.

For any two elements of $V^k$, $x = (x_1, x_2, \cdots, x_n)$ and $y = (y_1, y_2, \cdots, y_n)$, if $x_i \ge y_i (i = \overline{1, n})$, then we define as $x \ge y$.

Then the following properties are held.

$F$: fuzzy logic function, and $x, y \in V^h$!

$1°$ $x \in V_2^n \Rightarrow F(x) \in V_2$

$2°$ $x \ge y \Rightarrow F(x) \ge F(V)$

Property 1° shows that fuzzy logic function contains boolean logic function as a special case, and 2° shows that fuzzy logic function has monotonousness for fuzzy relation $\ge$.

(Example 1.2.4) Fuzzy logic function

$$F(x_1, x_2, x_3) = (x_1 \vee \bar{x}_2 \cdot x_3) \cdot x_2 \vee x_1 \cdot \bar{x}_2 \cdot \bar{x}_3$$

$$x = (0.2, 0.8, 0.6), y = (0.1, 0.9, 0.7), \quad x \ge y$$

$$F(x) = (0.2 \vee (1 - 0.8) \cdot 0.6) \cdot 0.8 \vee 0.2 \cdot (1 - 0.8) \cdot (1 - 0.6) =$$

$$(0.2 \vee 0.2 \cdot 0.6) \cdot 0.8 \vee 0.2 \cdot 0.2 \cdot 0.4 = 0.2 \cdot 0.8 \vee 0.2 \cdot 0.2 \cdot 0.4 = 0.2$$

$$F(y) = (0.1 \vee (1-0.9) \cdot 0.7) \cdot 0.9 \vee 0.1 \cdot (1-0.9) \cdot (1-0.7) =$$

$$(0.1 \vee 0.1 \cdot 0.7) \cdot 0.9 \vee 0.1 \cdot 0.1 \cdot 0.3 = 0.1 \cdot 0.9 \vee 0.1 \cdot 0.1 \cdot 0.3 = 0.1$$

Therefore, $F(x) \geq F(y)$.

### 1.2.2. Analysis of fuzzy logic function and composition

Analysis of fuzzy logic function is the problem that obtains the range fuzzy variables $x_1, x_2, \cdots, x_n$ has its value, with knowing the range fuzzy logic function $F(x_1, x_2, \cdots, x_n)$ has its value.

For consideration, divide the interval $[0, 1]$ into n classes as follows.

$$[0, 1] = \bigcup_{i=1}^{n} c_i, \ c_i \wedge c_j = \phi \ (i \neq j) \tag{1.2.5}$$

where

$$c_i = [a_{i-1}, a_i] \quad (i = 1, 2, \cdots, n-1)$$

$$c_n = [a_{n-1}, 1] \tag{1.2.6}$$

$$a_0 = 0$$

$$0 = a_0 < a_1 < a_2 < \cdots < a_{n-1} < a_n = 1$$

We can give the meaning to each class. For example, in case $n = 3$, if $x$ belongs to the third class, it means that $x$ belongs to the given set, and If $x$ belongs to the first class, it means it is not clear.

If fuzzy logic function satisfies the following equation for all variables $x_1, x_2, \cdots, x_n$, it can be said that fuzzy logic function $F$ belongs to the $j$ th class.

$$F(x_1, x_2, \cdots, x_n) \in c_j, \quad 1 \leq j \leq n \tag{1.2.7}$$

(Example 1.2.5) If fuzzy logic function $F(x,\ y,\ z) = x\bar{y}z + \bar{x}\,\bar{y} + \bar{x}y\bar{z}$ belongs to the second class in case $n = 2$, which value must $x,\ y,\ z$ have?

This problem is that find the range of $x,\ y,\ z$ to satisfy $a_1 \le F(x,\ y,\ z) \le 1$, according to the (1.2.7).

To satisfy $F(x,\ y,\ z) \ge a_1$ , $x\bar{y}z \ge a_1$ , $\bar{x}\,\bar{y} \ge a_1$ , or $\bar{x}y\bar{z} \ge a_1$ must be satisfied.

Next, obtain the range of $x,\ y,\ z$ for each item.

For $x\bar{y}z \ge a_1$, $x \ge a_1$, $\bar{y} \ge a_1$, $z \ge a_1$ must be satisfied.

$\bar{y} \ge a_1$ can be changed into $y \le 1 - a_1$. In the same method, we can handle the other items.

$$\begin{cases} x \ge a_1 \\ y \le 1 - a_1 \\ z \ge a_1 \end{cases} \quad \begin{cases} x \le 1 - a_1 \\ y \le 1 - a_1 \end{cases} \quad \begin{cases} x \le 1 - a_1 \\ y \ge a_1 \\ z \le 1 - a_1 \end{cases}$$

(Example 1.2.6) In general, obtain the range of $x$,$y$,$z$ so that fuzzy logic function $F(x,\ y,\ z) = \bar{x}\,\bar{y} + xy\bar{z}$ belongs to the j[th] class $c_j$.

Because $F \in c_j$ is equivalent to $a_{j-1} \le F < a_i$, to satisfy $F \in c_j$, $\bar{x}\,\bar{y} \ge a_{j-1}$, $xy\bar{z} \ge a_{j-1}$, $\bar{x}\,\bar{y} < a_j$ or $xy\bar{z} < a_j$ must be satisfied.

Solving this,

$$\begin{cases} x \le 1 - a_{j-1} \\ y \le 1 - a_{j-1} \end{cases} \quad \text{or} \quad \begin{cases} x \ge a_{j-1} \\ y \ge a_{j-1} \\ z \le 1 - a_{j-1} \end{cases}$$

and

$$\begin{cases} x > 1 - a_j \\ y > 1 - a_j \end{cases} \quad \text{and} \quad \begin{cases} x < a_j \\ y < a_j \\ z > 1 - a_j \end{cases}$$

As you can see in the example, the relation between the logic function and the inequality satisfying the class $c_j$ is as follows.

① Character related with $\geq$ is positive in *F* and character related with $\leq$ is negative in *F*.

② Character related with > is negative in *F* and character related with < is positive in *F*.

By using the above relation, given the fuzzy logic function, inequality can be obtained systemically as ① and ②.

(Example 1.2.7) Given the fuzzy logic function $F(x,y,z,u) = \bar{x}y\bar{z} + xy\bar{u} + x\bar{z}u$, find the condition *F* belongs to $c_j$.

By ①, for $a_{j-1}$,

$$\begin{cases} x \leq 1 - a_{j-1} \\ y \geq a_{j-1} \\ z \leq 1 - a_{j-1} \end{cases}$$

$$\text{or} \quad \begin{cases} x \geq a_{j-1} \\ y \geq a_{j-1} \\ u \leq 1 - a_{j-1} \end{cases}$$

$$\text{or} \quad \begin{cases} x \geq a_{j-1} \\ z \leq 1 - a_{j-1} \\ u \geq a_{j-1} \end{cases}$$

And also by ②, for $a_j$,

$$\begin{cases} x > 1 - a_j \\ y < a_j \\ z > 1 - a_j \end{cases} \begin{cases} x < a_j \\ y < a_j \\ u > 1 - a_j \end{cases} \begin{cases} x < a_j \\ z > 1 - a_j \\ u < a_j \end{cases}.$$

Next, consider the combination problem, the inverse problem of analysis problem, that is, to find the fuzzy logic function with knowing the range of variables to belong to class $c_j$.

(Example 1.2.8) When fuzzy functions $x, y, z$ are

$$\begin{cases} x \geq a_{n-1} \\ y \leq 1 - a_{n-1} \end{cases} \text{ or } \begin{cases} x \geq a_{n-1} \\ y \geq a_{n-1} \\ z \leq 1 - a_{n-1} \end{cases} \text{ or } \begin{cases} x \leq 1 - a_{n-1} \\ y \leq 1 - a_{n-1} \\ z \geq a_{n-1} \end{cases}$$

find the fuzzy logic function belonging to the class $c_n$.

Fuzzy logic function belonging to the class $c_n$ must satisfy the following inequality.

$$F(x, y, z) \geq a_{n-1}$$

Solving the above inequality inversely,

$$x\bar{y} \geq a_{n-1} \text{ or } xy\bar{z} \geq a_{n-1} \text{ or } \bar{x}\bar{y}z \geq a_{n-1}$$

So

$$F(x, y, z) = x\bar{y} + xy\bar{z} + \bar{x}\bar{y}z$$

(Example 1.2.9) Satisfied the fuzzy variables $x,y,z$ one of the following inequalities, find the fuzzy logic function $F(x, y, z)$ belonging the class $c_n$.

$$\left. \begin{array}{l} x \geq a_{n-1} \\ y \geq a_{n-1} \\ x \leq 1 - a_{n-1} \text{ or } z \geq a_{n-1} \end{array} \right\}$$

$$\left.\begin{array}{l} x \geq a_{n-1} \\ z \leq 1-a_{n-1} \end{array}\right\}$$

Solving the above inequalities inversely,

$$x \cdot y(\bar{x}+z) \geq a_{n-1} \text{ or } x\bar{z} \geq a_{n-1}$$

Therefore, fuzzy logic function to find is

$$F(x,\ y,\ z) = xy(\bar{x}+z) + x\bar{z}\ .$$

(Example 1.2.10) Satisfied the fuzzy variables $x,\ y,\ z$ one of the following inequalities, find the fuzzy logic function $F(x,\ y,\ z)$ belonging the class $c_n$.

$$\left.\begin{array}{l} x \leq t_1 \\ y \geq t_2 \end{array}\right\} \text{ or } \left.\begin{array}{l} x \geq t_3 \\ y \geq t_4 \\ z \leq t_5 \end{array}\right\} \qquad \text{(a)}$$

$$\left.\begin{array}{l} x > t_6 \\ y < t_7 \end{array}\right\} \text{ or } \left.\begin{array}{l} x < t_8 \\ x < t_9 \\ z > t_{10} \end{array}\right\} \qquad \text{(b)}$$

This problem is not related with the boundary values $a_{j-1}$ and $a_j$ of class $c_j$, but related with any value of $t_1,\ t_2,\ \cdots,\ t_{10}$ in $[0,\ 1]$ unlike the previous examples. But by multiplying a certain parameter $\omega_i (i=1,\ 2,\ \cdots,\ 10)$, it can be related with the boundary value of class $c_j$.

In this method, refer the fuzzy logic function for a) and b) to

$$F_1(x,\ y,\ z) = (\omega_1\bar{x}) \cdot (\omega_2 y) + (\omega_3 x) \cdot (\omega_4 y) \cdot (\omega_5 \bar{z})$$
$$F_2(x,\ y,\ z) = (\omega_6\bar{x}) \cdot (\omega_7 y) + (\omega_8 x) \cdot (\omega_9 y) \cdot (\omega_{10} \bar{z})$$

And then make the function so as to $F_1 \geq a_{j-1}$ and $F_2 < a_j$.

$\omega_i (i=1,\ 2,\ \cdots,\ 10)$ satisfying these conditions can be obtained as follows.

That is, in

$$\left.\begin{array}{l}\omega_1\bar{x} \ge a_{j-1} \\ \omega_2 y \ge a_{j-1}\end{array}\right\} \quad \text{or} \quad \left.\begin{array}{l}\omega_3 x \ge a_{j-1} \\ \omega_4 y \ge a_{j-1} \\ \omega_5 \bar{z} \ge a_{j-1}\end{array}\right\},$$

$$\bar{x} \ge \frac{a_{j-1}}{\omega_1} \Rightarrow x \le 1 - \frac{a_{j-1}}{\omega_1} = t_1, \ \therefore \omega_1 = \frac{a_{j-1}}{1-t_1}$$

$$y \ge \frac{a_{j-1}}{\omega_2} = t_2, \ \therefore \omega_2 = \frac{a_{j-1}}{t_2}$$

$$x \ge \frac{a_{j-1}}{\omega_3} = t_3, \ \therefore \omega_3 = \frac{a_{j-1}}{t_3}$$

$$y \ge \frac{a_{j-1}}{\omega_4} = t_4, \ \therefore \omega_4 = \frac{a_{j-1}}{t_4}$$

$$\bar{z} \ge \frac{a_{j-1}}{\omega_5} \Rightarrow z \le 1 - \frac{a_{j-1}}{\omega_5} = t_5, \ \therefore \omega_5 = \frac{a_{j-1}}{1-t_5}$$

and in

$$\left.\begin{array}{l}\omega_6\bar{x} < a_j \\ \omega_7 \bar{y} < a_j\end{array}\right\} \quad \text{or} \quad \left.\begin{array}{l}\omega_8 x < a_j \\ \omega_9 y < a_j \\ \omega_{10} \bar{z} < a_j\end{array}\right\},$$

$$\bar{x} < \frac{a_j}{\omega_6} \Rightarrow x > 1 - \frac{a_j}{\omega_0} = t_6, \ \therefore \omega_6 = \frac{a_j}{1-t_6}$$

$$y < \frac{a_j}{\omega_7} = t_7, \ \therefore \omega_7 = \frac{a_j}{t_7}$$

$$x < \frac{a_j}{\omega_8} = t_8, \ \therefore \omega_8 = \frac{a_j}{t_8}$$

$$y < \frac{a_j}{\omega_9} = t_9, \ \therefore \omega_9 = \frac{a_j}{t_9}$$

$$z < \frac{a_j}{\omega_{10}} \Rightarrow z > 1 - \frac{a_j}{\omega_{10}} = t_{10}, \ \therefore \omega_{10} = \frac{a_j}{1-t_{10}}.$$

Substituting such obtained $\omega_1, \omega_2, \cdots, \omega_{10}$ in $F_1$ and $F_2$, corresponding fuzzy logic function can be obtained.

## 1.3 Language Hedge

### 1.3.1 Set theoretical expression of language

It is an essential task to quantify and mathematize the language by giving the mathematical definition to some words to make the computer understand the natural language.

Word is a minimal unit of natural language and the word that is basic in words and is not decomposed is called original word. For example, word such as dog, horse, white, beautiful and so on are all original words.

Compound words can be made from words, and it was composed by applying the logic operations 《∧》, 《∨ and 《¬》 to the language in the view of mathematical logic. And operations 《∧》, 《∨ and 《¬》 are corresponding to 《∩》, 《∪》 and 《¬》 in set theory. So it is possible to make the compound word with words and decompose the compound word into words. Compound words can be expressed in the way of bracketing the words as follows.

White horse＝[horse]∩[white],

East north＝[east]∩[north],

brothers＝[elder brother]∪[younger brother],

desk and chair＝[desk]∪[chair],

nonmetal＝¬[metal]

Hedge of language can be done by introducing the concept of operator on the basis of set theoretical expression of language. For example, we can consider 《extremely》, 《very》, 《comparatively》, 《quite》, 《more or less》, 《almost》,

《probably》 and so on as operators.

## 1.3.2. Language variable

Language variable is the generalized concept of fuzzy variable. Fuzzy variable is characterized as the three dimensional pair $(X,\ U,\ R(X\!:u))$, where $X$ is the name of variable, $U$ is the set of objects, and $R(X\!:u)$ is the restriction. Restriction $R(X\!:u)$ is the fuzzy set in $U$ determined by fuzzy variable $X$. If $R(X:u)$ is general set, then it becomes the variable that has been used in mathematics. Like this, fuzzy variable is the generalization of variable.

Language variable is denoted as $(X,\ U,\ R(X:u))$, where restriction is the whole of various fuzzy subsets in object set.

For example, consider the adjective 《pretty》. This is the word that indicates the external characteristics of a certain object. This is considered as an expression of fuzzy set restricting the fuzzy variable named 《pretty》. Applying rhetoric such as 《very》, 《not》, 《extremely》, 《quite》 and so on to the fuzzy set 《pretty》, new fuzzy sets such as 《very pretty》, 《extremely pretty》, 《not pretty》 and 《quite pretty》. Such fuzzy sets are the values of language variables such as fuzzy set 《pretty》.

A main characteristic of language variable is that language variable selects the fuzzy variable as its value.

For example, language variable 《age》, expressing the age, can have the following values. 《young》, 《not young》, 《old》, 《very old》, 《not young and not old》 and 《quite old》. Here 《young》 is the name of fuzzy variable reflecting the meaning 《young》, and 《old》 is meaning 《old》.

Referring $X$ to the name of fuzzy variable, restriction decided by $X$ is the meaning of $X$.

For example, restriction decided by fuzzy variable 《old》 is the fuzzy set defined in $U=[0,\ 100]$ like (1.3.1).

$$R(\text{old})=\int_{50}^{100}\left[1+\left(\frac{u-50}{5}\right)^{-2}\right]^{-1}\Big/u,\ u\in U \tag{1.3.1}$$

Fuzzy set expressed by $R(\text{old})$ can be considered that it has the meaning 《old》.

The other characteristic of language variable is that it is combined according to the following two rules.

① Syntactic rule: this means the algorithm for generating the name of value of variable.

② Semantic rule: this is the rule to decide the algorithm for calculating the meaning of each value.

These rules are the essential parts reflecting the characteristics of structural variable.

Language variable is characterized as 5 dimensional pair as follows.

$$(N,\ T(N),\ U,\ G,\ M) \tag{1.3.2}$$

In (1.3.2), $N$ is the name of variable, and $T(N)$ is the noun set of $N$, that is, the syntactic rule for generating the value name $X$ of $N$.

$M$ is the semantic rule for expressing the meaning of $X$ as the fuzzy set in $U$.

Especially, $X$ that is the name generated by grammar $G$ is called noun.

Noun that has a function, composed with one or more words is called basic noun, and noun consisted of one or more basic words is called compound noun.

Referring $X_1,\ X_2,\ \cdots$ to the nouns of $T$, $T$ is denoted as follows.

$$T = X_1 + X_2 + \cdots$$

Where $T$ may be expressed as $T(G)$ when it is necessary that $T$ was generated by grammar $G$. The meaning of noun $X$ $M(X)$ is defined as (1.3.3) so as to be the restriction $R(X)$ decided by fuzzy variable $X$.

$$M(x) \triangleq R(X) \tag{1.3.3}$$

This means that $R(X)$ is the fuzzy set representing the meaning of $X$.

(Example 1.3.1) Consider the language variable 《Age》.

Language value of 《Age》 may be 《old》 and 《old》 is the basic noun. And it has the other value 《very old》, which is compound noun. Besides this, there are many elements of noun set of 《Age》.

$T$(Age)=old + very old + not old + more or less young + quite young + not very old and not very young+ …

where each noun is the name of fuzzy variable in $U = [0,\ 100]$. Then the meaning of 《old》 is given as (1.3.4).

$$M(\text{old}) = \int_{50}^{100} \left[ 1 + \left( \frac{u-50}{5} \right)^{-2} \right]^{-1} \Big/ u \tag{1.3.4}$$

Or simply,

$$\text{old} = \int_{50}^{100} \left[ 1 + \left( \frac{u-50}{5} \right)^{-2} \right]^{-1} \Big/ u \qquad (1.3.5)$$

Similarly, the meaning of language value of 《very old》 can be described as follows.

$$M(\text{veryold}) = \text{veryold} = \int_{50}^{100} \left[ 1 + \left( \frac{u-50}{5} \right)^{-2} \right]^{-2} \Big/ u \qquad (1.3.6)$$

(Example 1.3.2) Consider the language variable 《Number》. Language variable is as follows.

$$T(\text{Number}) = \text{few} + \text{several} + \text{many}$$

$$U = 1 + 2 + \cdots + 10$$

where 《few》, 《several》 and 《many》 are fuzzy variables defined in $U = 1 + 2 + \cdots + 10$.

$$\text{few} = 0.4/1 + 0.8/2 + 1/3 + 0.4/4$$

$$\text{several} = 0.5/3 + 0.8/4 + 1/5 + 1/6 + 0.8/7 + 0.5/8,$$

$$\text{many} = 0.4/6 + 0.6/7 + 0.8/8 + 0.9/9 + 1/10$$

That language variable 《Number》 has 《few》 as its value is denoted as follows.

$$\text{Number} = \text{few}$$

Next, consider the structural language variable.

When given language variable $N$, its noun set $T(N)$ and $M$ is characterized algorithmically, the language variable $N$ is called structural language variable.

In this sense, syntactic rule combined with the structural language variable can be considered as the algorithm for calculating the meaning of each noun in $T(N)$.

(Example 1.3.3) Consider the role of syntactic and semantic rules of structural language variable using the language variable 《Age》.

Nouns of 《Age》 are 《old》, 《very old》, 《very very old》. Therefore, noun set of 《Age》 can be denoted as follows.

$$T(\text{Age})=\text{old} + \text{very old} + \text{very very old} + \ldots$$

It is certain that in the simple case, each noun of $T(\text{Age})$ is one of 《old》 or 《very old》.

Induct this rule in general way.

Denote the connection of $x$ and $y$ as $x \cdot y$. If $x = \text{very},\ y = \text{old}$, then $xy = \text{very old}$.

If $A,\ B$ are the sets of rows, that is

$$A = x_1 + x_2 + \cdots$$
$$B = y_1 + y_2 + \cdots$$

, then connection of $A$ and $B$ is denoted as $AB$, and is defined as the following set.

$$AB = (x_1 + x_2 + \cdots)(y_1 + y_2 + \cdots) = \sum_{ij} x_i y_j$$

For example, if $A = \text{very},\ B = \text{old} + \text{very old}$, then

$$AB = \text{very(old} + \text{very old)} = \text{very old} + \text{very very old}$$

Using such expression method, we can make $T(\text{Age})$ the solution of the following equation.

$$T=\text{old} + \text{very } T \qquad (1.3.7)$$

(1.3.7) can be solved using the following recursion equation.

$$T^{i+1} = \text{old} + \text{very} T^i,\ i = 0,\ 1,\ 2,\ \cdots \qquad (1.3.8)$$

Setting the initial value of $T^i$ to empty set 0,

$$\begin{aligned} &T^0 = 0 \\ &T^1 = \text{old} \\ &T^2 = \text{old} + \text{very old} \qquad (1.3.9) \\ &T^3 = \text{old} + \text{very old} + \text{very very old} \\ &\cdots\ \cdots \end{aligned}$$

As result, solution of equation is as follows.

$$T = T^\infty = \text{old} + \text{very old} + \text{very very old} + \cdots \qquad (1.3.10)$$

In this example, syntactic rule is expressed by (1.3.7) and its solution is obtained as follows.

Considering that (1.3.7) is algebraic expression, syntactic rule can be expressed as the following generation system.

$$T \to \text{old} \qquad (1.3.11)$$

$$T \to \text{very old} \qquad (1.3.12)$$

In this case, noun of $T$ is obtained using the above generation rule. That is, if $T$ is very $T$ and, if $T$ is old in very $T$, then noun very old is obtained.

$$T \to \text{very}\,T \to \text{very very}\,T \to \text{very very very}\,T \to \text{very very very old}. \qquad (1.3.13)$$

Next, consider the semantic rules for 《Age》.

To compute the meaning of noun such as very … very old, first it is necessary to know the meaning old and very, where noun old is the initial value. That is, to compute the meaning of compound noun in $T$, at least the meaning of old must be characterized as the initial data.

Noun very has the role of modification. Suppose that very has the rule of concentration operator, the approximation of very is expressed as follows.

$$\text{very old} = \text{CON(old)} = \text{old}^2 \tag{1.3.14}$$

As result, semantic rule of Age is expressed as follows.

$$M\,(\text{very} \ldots \text{very old}) = \text{old}^{2n}$$

where $n$ is the number of very in very … very old. If basic noun old is

$$\text{old} = \int_{50}^{100} \left[1 + \left(\frac{u-50}{5}\right)^{-2}\right]^{-1} \bigg/ u \tag{1.3.15}$$

, then

$$M(\text{very} \ldots \text{very old}) = \int_{50}^{100} \left[1 + \left(\frac{u-50}{5}\right)^{-2}\right]^{-2n} \bigg/ u\,, \quad n = 1,\ 2, \cdots \tag{1.3.16}$$

This equation offers the semantic rule for computing the meaning of compound noun that basic noun old is generated by equation $T$=old + very $T$ by modifier very.

Next, consider the boolean language variable used very conveniently in application.

Boolean language variable is the following type of language variable. That is, $X$ is the boolean expression of variables that has the type of $X_p$, $hX_p$ or $hX$, where $h$ is language modifier, $X_p$ is noun, and $hX$ is the name of fuzzy set obtained by applying $h$ to $X$.

not very young and not very old

For example, in the above expression, $h$ is very, $X_p$ is young and old. Similarly, in very very young, $h$ is very very, and $X_p$ is young.

(Example 1.3.4) Refer Age to the boolean language variable having the following noun set.

*T*(Age)=young + not young +old +not old + very young + not young and not old + young or old + young or( very young and not very old ) + …

Then, the meaning of typical value of Age can be described as follows.

*M* (not young) = *T*(young),

*M* (not very young) = *T* ($young^2$),

*M* (not very young and not very old) = *T* ($young^2$) ∩ *T* ($old^2$),

*M* (young or old) = *T*(young ∪ old)

These expressions show that the meaning of compound noun is expressed by the meaning of basic nouns. So If young and old are as follows,

$$\text{young} = \int_0^{25} 1/u + \int_{25}^{100} \left[1 + \left(\frac{u-25}{5}\right)^{-2}\right]^{-1} \Big/ u \qquad (1.3.17)$$

$$\text{old} = \int_{50}^{100} \left[1 + \left(\frac{u-50}{5}\right)^{-2}\right]^{-1} \Big/ u \qquad (1.3.18)$$

Then

$$M(\text{young or old}) = \int_0^{25} 1/u + \int_{25}^{50} \left[1 + \left(\frac{u-25}{5}\right)^{-2}\right]^{-1} \Big/ u +$$

$$+ \int_{50}^{100} \left[1 + \left(\frac{u-25}{5}\right)^{-2}\right]^{-1} \vee \left[1 + \left(\frac{u-50}{5}\right)^{-2}\right]^{-1} \Big/ u \qquad (1.3.19)$$

## 1.3.3 Language Hedge

We explain the language hedge as the operator converting one fuzzy set such as 《very》 to the other fuzzy set for example modifier named as much, more or less, rather.

To discuss the language hedge, we must consider the operations about fuzzy set, which is square operation, concentration, dilatation, contrast, standardization, fuzzification and so on. We describe these operations.

[Definition 1.3.1] Referring fuzzy set in $Y$ to $A$, and $\alpha$ to real number, square $A^{\alpha}$ as the expansion of algebraic product for fuzzy set is defined as (1.3.20).

$$A^{\alpha} \Leftrightarrow \mu_{A^{\alpha}}(y) = \mu_A(y)^{\alpha} \tag{1.3.20}$$

And If $\alpha$ is positive number, scalar product $\alpha A$ is defined as follows.

$$\alpha A \Leftrightarrow \mu_{\alpha A}(y) = \alpha \cdot \mu_A(y)$$

(Example 1.3.5) If $A = 0.64/1 + 0.25/2 + 0.81/3$, then $A^2, A^{0.5}, 0.5A$ are calculates as follows.

$$\begin{aligned} A^2 &= 0.41/1 + 0.06/2 + 0.66/3, \\ A^{0.5} &= 0.8/1 + 0.5/2 + 0.9/3, \\ 0.5A &= 0.32/1 + 0.125/2 + 0.41/3 \end{aligned}$$

[Definition 1.3.2] Concentration CON($A$) for fuzzy set $A$ is defined as follows.

$$\mu_{\mathrm{CON}}(y) = \mu_A(y)^2 \tag{1.3.21}$$

This means CON($A$) $= A^2$.

[Definition 1.3.3] Dilatation for fuzzy set $A$ is denoted as DIL($A$), and is defined as follows.

$$\mathrm{DIL}(A) = A^{0.5} \tag{1.3.22}$$

Namely,

$$\mu_{\mathrm{DIL}(A)}(y) = \mu_A(y)^{0.5} = \sqrt{\mu_A(y)} \tag{1.3.23}$$

[Definition 1.3.4] Contrast for fuzzy set $A$, and is defined as follows.

$$\mathrm{INT}(A) = \begin{cases} 2A^2, & 0 \le \mu_A(y) \le 0.5 \\ 1-2(1-A)^2, & 0.5 < \mu_A(y) \le 1 \end{cases} \tag{1.3.24}$$

Namely,

$$\mu_{\mathrm{INT}(A)}(y) = \begin{cases} 2\mu_A(y)^2, & 0 \le \mu_A(y) \le 0.5 \\ 1-2(1-\mu_A(y))^2, & 0.5 < \mu_A(y) \le 1 \end{cases} \tag{1.3.25}$$

CON and INT satisfy the following property.

$$\mathrm{CON}(A \cup B) = \mathrm{CON}(A) \cup \mathrm{CON}(B) \tag{1.3.26}$$

$$\mathrm{CON}(A \cap B) = \mathrm{CON}(A) \cap \mathrm{CON}(B) \tag{1.3.27}$$

$$\mathrm{CON}(AB) = \mathrm{CON}(A)\mathrm{CON}(B) \tag{1.3.28}$$

$$\mathrm{INT}(A \cup B) = \mathrm{INT}(A) \cup \mathrm{INT}(B) \tag{1.3.29}$$

$$\mathrm{INT}(A \cap B) = \mathrm{INT}(A) \cap \mathrm{INT}(B) \tag{1.3.30}$$

$$\mathrm{INT}(AB) = \mathrm{INT}(A)\mathrm{INT}(B) \tag{1.3.31}$$

(Example 1.3.6) If $A = 1/1 + 0.8/2 + 0.6/3 + 0.4/4 + 0.2/5$, then

$$\mathrm{CON}(A) = 1/1 + 0.64/2 + 0.36/3 + 0.16/4 + 0.04/5$$

$$\mathrm{DIL}(A) = 1/1 + 0.89/2 + 0.77/3 + 0.63/4 + 0.45/5$$

$$\mathrm{INT}(A) = 1/1 + 0.92/2 + 0.68/3 + 0.32/4 + 0.08/5$$

Referring the maximum membership of membership function $\mu_A$ of fuzzy set $A$ in $Y$,

$$\bar{\mu}_A = \vee_y \mu_A(y)$$

If $\bar{\mu}_A = 1$, then $A$ is standard fuzzy set. If $\bar{\mu}_A \neq 1$, $\mu_A$ can be made to standard fuzzy set, by dividing $\mu_A$ into $\bar{\mu}_A$. This operation is called standardization, and is denoted as NORM($A$). That is,

$$\mathrm{NOR}M(A) = \frac{1}{\mu_A} A \tag{1.3.32}$$

(Example 1.3.7) If $A = 0.8/y_1 + 0.4/y_2 + 0.6/y_3 + 0.2/y_4$, then A is standard set and it is as follows, because $\bar{\mu}_A = 0.8$.

$$\mathrm{NOR}M(A) = 1/y_1 + 0.5/y_2 + 0.75/y_3 + 0.25/y_4$$

Fuzzification has effectiveness converting the crisp set to fuzzy set and increasing the degree of fuzzy.

[Definition 1.3.5] Denoting fuzzification operator as $F$, the result applied $F$ to fuzzy set $A$ is the following fuzzy set $F(A:K)$. That is,

$$F(A:K) = \sum_i \mu_A(y_i)K(y_i) \tag{1.3.33}$$

where $K(y_i)$ is the fuzzy set in $Y$ called as the kernel of $F$. Kernel $K(y_i)$ expresses the fuzzy set $y_i$ that is approximately equivalent to $y_i$. And $\mu_A(y_i)K(y_i)$ expresses the algebraic product of scalar $\mu_A(y_i)$ and fuzzy set $K(y_i)$.

(Example 1.3.8) If

$$Y = \{1,2,3,4\}$$

$$A = 0.8/1 + 0.6/2$$

$$K(1)=1/1+0.4/2, K(2)=0.4/1+1/2+0.4/3,$$

then $F(A{:}K)$ is as follows.

$$\begin{aligned} F(A{:}K) &= \mu_A(1)\cdot K(1)+\mu_A(2)K(2)= \\ &=0.8(1/1+0.4/2)+0.6(0.4/1+1/2+0.4/3)= \\ &=0.8/1+0.32/2+0.24/1+0.6/2+0.24/3= \\ &=0.8/1+0.6/2+0.24/3 \end{aligned}$$

Using these operations, we explain various language modifications. Language modification

[very]

Referring X to noun, very is expresses as follows, as we considered before.

$$\text{very}X = X^2$$

That is, if

$$X=\mu_1/y_1+\mu_2/y_2+\cdots+\mu_n/y_n$$

, then

$$\text{very } X=\mu_1^2/y_1+\mu_2^2/y_2+\cdots+\mu_n^2/y_n$$

[plus and minus]

Plus and minus are the artificial language modification and operators acting more slowly than concentration and dilatation.

$$\text{plus}X = X^{1.25}$$
$$\text{minus}X = X^{0.75}$$

where Exponent of $X$ 1.25 and 0.75 are the number selects so as to satisfy the following equation.

$$\text{plus plus } X = \text{minus very } X$$

By using plus and minus, highly is defined as follows.

highly=minus very very =plus plus very

[more or less]

More or less is more or less intelligent, more or less rectangular,

More or less is used like more or less intelligent, more or less rectangular, more or less sweet and so on.

More or less play a role of fuzzification. For example, if fuzzy variable 《recent》 is expressed as recent, that is

$$\text{recent} = 1 / 2003+0.9 / 2002 + 0.7 / 2001 + 0.5 / 2000$$

, then

$$\text{more or less recent} = F ( \text{recent} : K )$$

and if kernel $K$ is

$$\begin{aligned} K(2003) &= 1/2003 + 0.9/2002 \\ K(2002) &= 1/2002 + 0.8/2001 \\ K(2001) &= 1/2001 + 0.8/2000 \\ K(2000) &= 1/2001 + 0.8/1999 \end{aligned}$$

, then more or less recent =1/2 003+0.9/2 002+0.72/2 001+0.56/2 000+0.4/1 999

[sort of]

Sort of has the effectiveness that move the degree of membership of the centre object (its degree of membership is near 1) of $X$ lower, and move the degree of object which fuzzy degree is high higher.

Sort of is defined as follows.

sort of $X$=NORM $(\neg\text{COW}(x^2) \cap \text{DIL}(x))$

sort of $X$=NORM $(\text{INT}(\text{DIL}(x)) \cap \text{INT}(\text{DIL}(\neg x)))$

[rather]

Rather is expressed as follows.

rather $X$ = NORM ( INT ( $x$ ) )

rather $X$ = NORM ( INT ( COW ( $x$ ) ) $\cap \neg$ ( CON ( $x$ ) ) )

So far, we considered several kinds of language modification.

Applying the fuzzy variable to this language hedge, we can get effectiveness of language hedge.

# 1.4 Fuzzy Language and Fuzzy Grammar

## 1.4.1 Fuzzy language

In order that the computer simulates the human's thinking better, natural language must be combined with the current formulated algorithm language. By inserting the natural language to the algorithm language, computer will be able to deal with the fuzzy concept and serve better for the human, comprehending the human intelligence. Such a computer language is called fuzzy algorithm language.

To make a computer understand and process the fuzzy concept, we must select the object space first of all.

In general, we select the set of real number or its subset as the object space and define basic word such as 《small》, 《large》, 《long》, 《short》, 《heavy》 and 《light》 on the object space. And then by adding the fuzzy modification operators, we obtain 《very large》, 《quite light》, 《more or less small》 and so on, and adding fuzzy set operators 《∨》 and 《∧》, obtain 《neither large nor small》, 《larger or comparatively large》 and so on.

Programming these concepts, we can obtain value setting statements and program the fuzzy condition statements.

Sometimes, the result computed using the fuzzy algorithm language may be fuzzy, and in this case, it is possible to convert the fuzzy result to the definite one, by using the $\lambda$ level set.

We consider the fuzzy algorithm language FSTDS (Fuzzy Set Theory Data System).

There are 52 kinds of operation symbols in FSTDS and they can be divided into 8 kinds.

① Operation symbol configuring the set

There are 《Set》 and 《Fset》.

《Set》 configures the general set and 《Fset》 does fuzzy set. For example, $\mathrm{Set}(a,b,c)$ represents the set $\{a,b,c\}$ and $\mathrm{Fset}\left(\frac{0.1}{a}, \frac{0.8}{c}\right)$ represents the set $A = \frac{0.1}{a} + \frac{0}{b} + \frac{0.8}{c}$.

② Assignment operation symbol.

This is the operation symbol named fuzzy set, and there is assignment symbol 《:=》, it is also written as 《Assign》.

③ Operation symbol on the same kind of fuzzy set

《Union》, 《Intersection》, 《Prod》 (product), 《Asum》 (algebraic sum), 《Bsum》 (boundary sum), 《Bdif》 (boundary difference)

④ Operation symbol on fuzzy relation

《Compose》 (composition), 《Converse》 (converse), 《Image》 (image), 》Domain》 (domain)

⑤ Relational operation symbol

《EQ》 (equal), 《Subset》 (subset), 《Element》 (element)

⑥ Other operation symbol

《Cut》(select the $\lambda$-level set), 《EXP》(exponent), 《#》(number of elements), 《Dlt》 (delete)

By using these operation symbols, we expand the function of FSTDS.

⑦ Output operation symbol

《Print》(standard output), 《Printb》(bool output), 《Prints》(output in the form of general set), 《Printn》 (output of set. Output the name of set. In case of no set name, output ***), 《Printc》 (only output the character symbol in the brackets)

⑧ Operation symbol adjusting the control type

《End》(conclusion), 《Dump》(delete the memory domain of operation object), 《Snap》 (output the configured fuzzy set and the output format is Print), 《Para》 (offers the various useful information in FSTDS)

Program design of FSTDS language

Given object space $X = \{a,\ b,\ c\}$, and fuzzy set on $X$ satisfies

$$A = 1/a + 0.9/b + 0.3/c,$$
$$B = 0.1/a + 0.7/b + 0.9/c$$

and fuzzy relation $R$ is the subset of $X \times X$

$$R = \begin{bmatrix} 1 & 0.8 & 0 \\ 0.7 & 1 & 0.2 \\ 0 & 0.5 & 0.1 \end{bmatrix}$$

, prove the following equation.

$$(A \cup B) \circ R = (A \circ R) \cup (B \circ R)$$

Program by FSTDS is as follows.

① $A$:= Fset$(1/a, 0.9/b, 0.3/c)$

② $B$:= Fset$(0.1/a, 0.7/b, 0.9/c)$

③ $R$:= Fset(1/[$a$, $a$], 0.8/[$a$, $b$], 0.7/[$b$, $a$], 1/[$b$, $b$], 0.2/[$b$, $c$],

0.5/[$c$, $b$], 0.1/[$c$, $c$])

④ Print(Assign(C, Union($A$, $B$)))

⑤ Print(Image($R$, $C$))

⑥ $D$:= Image($R$, $A$)

⑦ $E$:= Image($R$, $B$)

⑧ Print(Union($D$, $E$))

⑨ END

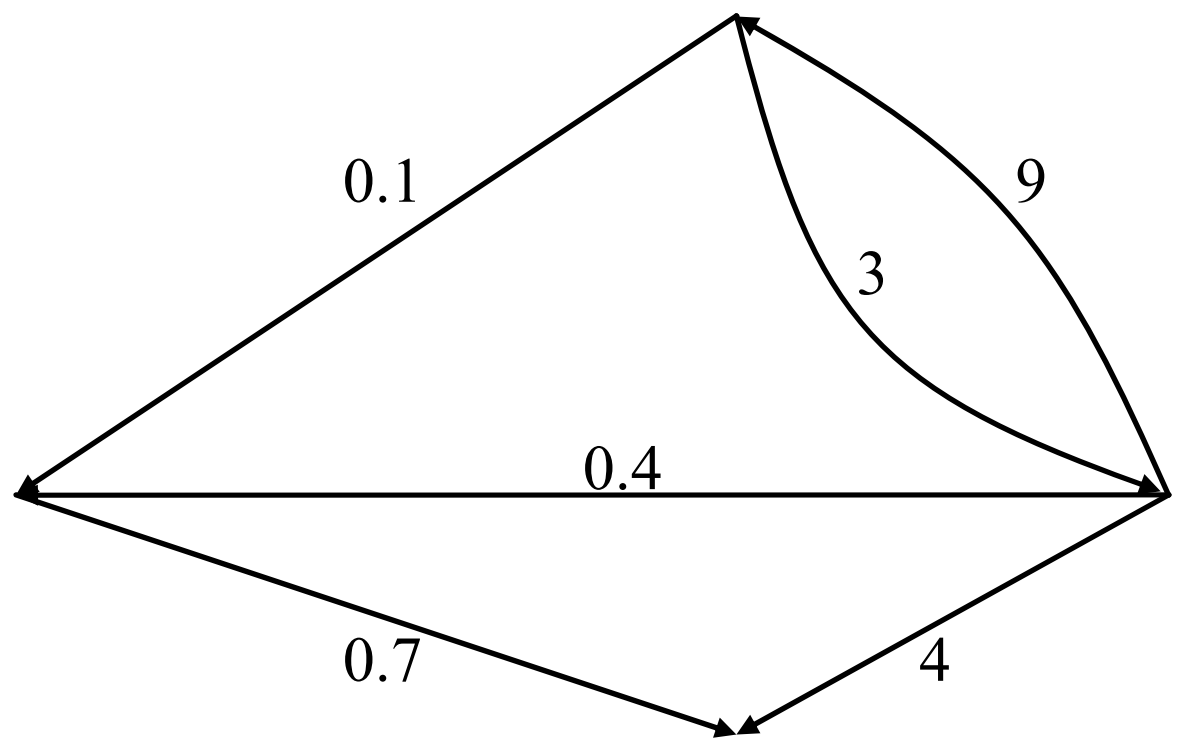

Figure 1.4.1 Fuzzy directed graph

In this program, among three output statements ④, ⑤, ⑧, Fset $(1/a, 0.9/b, 0.9/c)$ is corresponded to $A \cup B$ and Fset $(1/a, 0.3/b, 0.2/c)$ is to $(A \cup B) \cdot R$, and Fset $(1/a, 0.9/b, 0.2/c)$ is to $(A \circ R) \cup (B \circ R)$.

Next, FSTDS language program to input the fuzzy directed graph such as figure

1.4.1 can be written as follows.

$V$:=Set($x$, $y$, $z$, $w$)

$A$:= Fset(0.1/$\langle x, y\rangle$, 0.7/$\langle y, z\rangle$,

0.4/$\langle w, z\rangle$, 0.4/$\langle w, y\rangle$,0.3/$\langle x, w\rangle$, 0.9/$\langle w, x\rangle$)

$G$:= Fset($\langle V, A\rangle$)

## 1.4.2 Fuzzy grammar

Before consideration of fuzzy grammar, we simply consider the general grammar.

[Definition 1.4.1] General grammar $G$ is the set of the following 4 elements,

$$G = (V_T, V_N, S, P) \qquad (1.4.1)$$

where $V_T$: set of terminal symbols, $V_N$: set of nonterminal symbols, $S$: start symbol, $P$: set of rules.

$$P = \{(\alpha \to \beta) | \alpha, \beta \in V_T \cup V_N\} \qquad (1.4.2)$$

For example, grammar representing the isosceles triangle is

$$G = (V_T, V_N, S, P) \qquad (1.4.3)$$

Here $V_T = \{a, b, c\}$ : three sides of triangle, $V_N = \{A, B\}$; $A, B$ : nonterminal symbols

$$P = \{S \to aSc, A \to aAc, A \to bB, B \to bB, B \to b\}$$

According to this grammar, we can make some sentence representing the isosceles triangle. The generation method is as figure 1.4.2, and by selecting the terminal symbol of this grammar tree one by one, a sentence *aabbbcc* is generated.

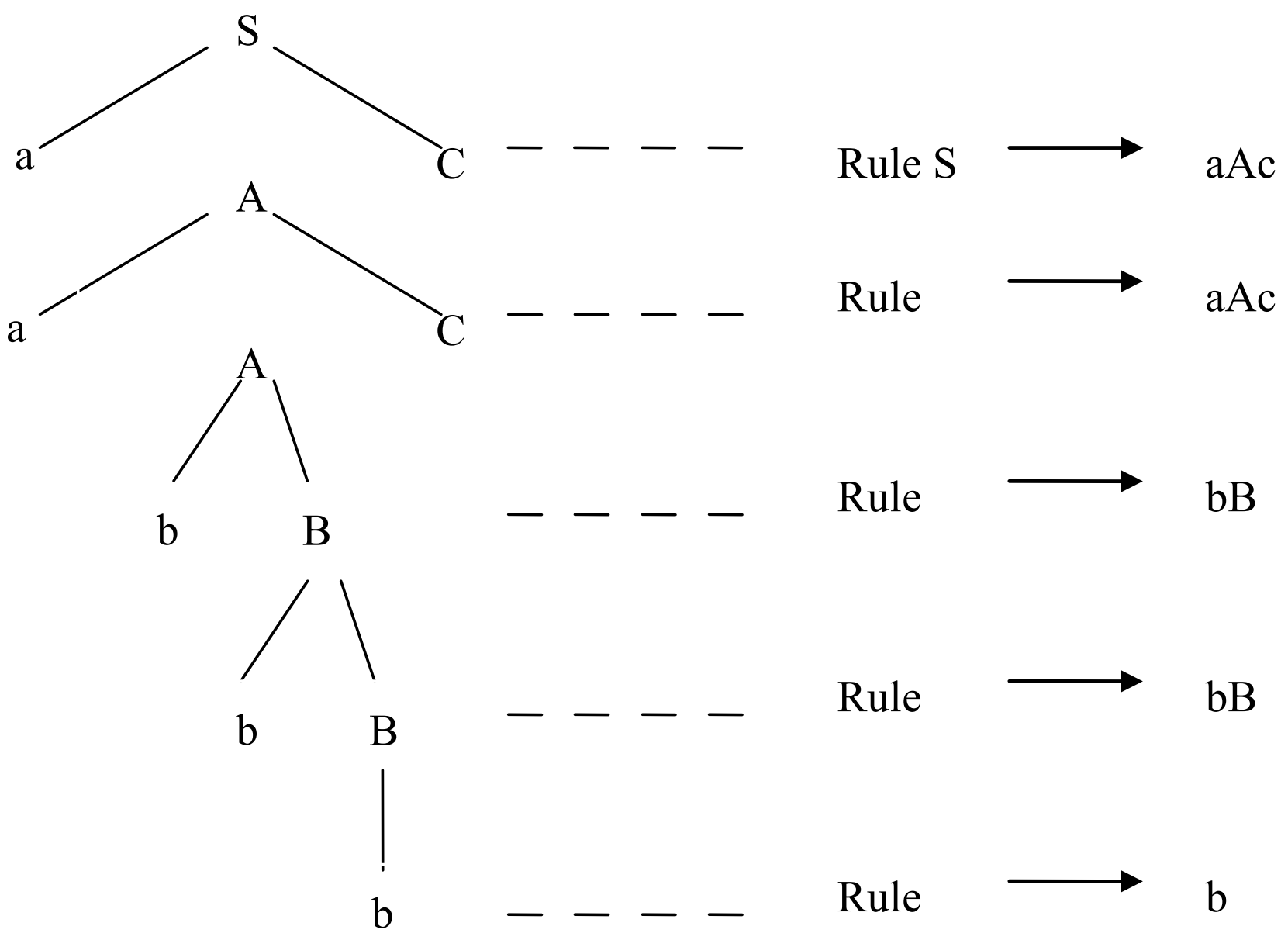

Figure 1.4.2 Tree-type structure of isosceles triangle grammar.

If a character expresses the length of corresponding side of triangle, above *aabbbcc* expresses that $a:b:c$ is 2:3:2.

No matter how we use rule *P* in this grammar, the number of last obtained *a* sand *c* are always the same. Therefore, the sentence generated by this grammar represents a isosceles triangle.

Grammar of fuzzy language makes the general grammar fuzzy and the generated sentence is

[Definition 1.4.2] Fuzzy grammar $G_F$ is the following set of 5 elements. That is,

$$G_F = (V_T,\ V_N,\ S,\ P,\ f) \qquad (1.4.4)$$

Here, $V_T$: set of terminal symbols, $V_N$: set of nonterminal symbols, $S$: start symbol, $P$: rule set, $f$: the following mapping

$$f{:}P \to [0,\ 1] \qquad (1.4.5)$$

Explain for example.

$$V_T = \{a,\ b\}$$
$$V_N = \{A,\ B,\ C\}$$

$S$: star symbol

$P$: the rule as follows.

$$P_1{:}S \to AB$$
$$P_2{:}A \to a$$
$$P_3{:}B \to b$$
$$P_4{:}A \to aAB$$
$$P_5{:}A \to aB$$
$$P_6{:}A \to aC$$
$$P_7{:}C \to a$$
$$P_8{:}C \to aa$$
$$P_9{:}A \to B$$

$F$ is the following mapping.

$$f(P_1) = 1$$
$$f(P_2) = 1$$
$$f(P_3) = 1$$
$$f(P_4) = 0.9$$
$$f(P_5) = 0.5$$
$$f(P_6) = 0.5$$
$$f(P_7) = 0.5$$
$$f(P_8) = 0.2$$
$$f(P_9) = 0.2$$

This is a fuzzy grammar and the sentences generated by this grammar are the fuzzy subset $A$ of $V_T^*$, where each sentence has the certain degree of membership for $A$. For example, If

$$S \xrightarrow[P_1]{1} AB \xrightarrow[P_4]{0.9} aABB \xrightarrow[P_4]{0.9} aaABBB$$

$$\xrightarrow[P_2]{1} aaaBBB \xrightarrow[P_3]{1} aaabBB \xrightarrow[P_3]{1} aaabbB \xrightarrow[p_3]{1} aaabbb$$

, then

$A(aaabbb) = f(P_1) \wedge f(P_4) \wedge f(P_4) \wedge f(P_2) \wedge f(P_3) \wedge f(P_3) \wedge f(P_3) =$

$= 1 \wedge 0.9 \wedge 0.9 \wedge 1 \wedge 1 \wedge 1 \wedge 1 = 0.9$

If

$$\underbrace{aa\cdots a}_{n}\underbrace{bb\cdots b}_{m}$$

is written simply as $a^n b^m$, we can prove the following.

If $n = m = 1$, then $A(a^n b^m) = 1$

If $n = m \neq 1$, then $A(a^n b^m) = 0.9$

If $m = n \pm 1$, then $A(a^n b^m) = 0.5$

If $m = n \pm 2$, then $A(a^n b^m) = 0.2$

Because $A(a) = 0$ is satisfied for any sentence $a \in V_T$, sentence generated by this fuzzy grammar is only the type of $a^n b^m$, and satisfies $|m - n| \leq 2$. If $a$, $b$ denote the length of two sides of triangle, then the sentence generated by this grammar represents a fuzzy model. That is, it expresses an approximate isosceles triangle.

Fuzzy grammar can be classified into four classes 0, Ⅰ, Ⅱ, Ⅲ as the general grammar, and they are corresponded to a certain automatic machine.

In fuzzy language, consider the application example of fuzzy grammar. Given the set of words

$$V = \{\text{young},\ \text{old}\}$$

and added combination operators 《and》, concentration 《very》 and negative 《no》,

set of terminal symbols is obtained. Among the set of rows corresponding to $V_T$, the following words have their meaning.

Words such as 《young》, 《old》, 《not young》, 《not old》, 《very young》, 《very old》, 《quite old》, 《quite young》, 《not quite young》, 《neither quite young nor quite old》, 《young and not old》, and so on can be generated by fuzzy grammar.

Consider the set of nonterminal symbols

$$V_N = \{A,\ B,\ C,\ O,\ Y\} \tag{1.4.6}$$

*S* is start symbol, and *P* is the rule of word formation. To avoid the confusion, suppose as follows. In the left bottom of word formation rule is tagged *L* and in the right is tagged *R*. for example, we can write the membership function for $S \to A$ as follows.

$$\mu(S_L) = (A_R) \tag{1.4.7}$$

Next, by using the following word formation rules for 《neither quite young nor old》,

$S \to A$, membership function $\mu(S_L) = \mu(A_R)$

$A \to B$, membership function $\mu(A_L) = \mu(B_R)$

$B \to C$, membership function $\mu(B_L) = \mu(C_R)$

$S \to S$ or $A$, membership function $\mu(S_L) = \mu(S_R) \vee \mu(A_R)$

$A \to A$ and $B$, membership function $\mu(A_L) = \mu(A_R) \wedge \mu(B_R)$

$B \to C$ no, membership function $\mu(B_L) = 1 - \mu(C_R)$

$0 \rightarrow$ very 0, membership function $\mu(Y_L) = \mu^2(Y_R)$

$C \rightarrow 0$, membership function $\mu(C_L) = \mu(O_R)$ $\mu(0_L) = \mu^2(0_R)$

$Y \rightarrow$ quite $Y$ membership function

$C \rightarrow Y$, membership function $\mu(C_L) = \mu(Y_R)$

$0 \rightarrow$ old, membership function $\mu(0_L) = \mu$ (old)

$Y \rightarrow$ young, membership function $\mu(Y_L) = \mu$ (young)

We can derive the followings.

$S \rightarrow A$

$\rightarrow$ $A$ and $B$

$\rightarrow$ not $C$ and $B$

$\rightarrow$ not very $Y$ and $B$

$\rightarrow$ not very young and $B$

$\rightarrow$ not very young and not $C$

$\rightarrow$ not very young and not $O$

$\rightarrow$ not very young and not very very $C$

$\rightarrow$ not very young and not very old

To derive these membership functions, we must decide the basic words 《old》, and 《young》 first of all. We have already been aware of their membership functions.

Then, derive according to the grammar tree of figure 1.4.3, the terminal symbol of the tree forms a sentence 《neither quite young nor quite old》. Lower tag of the nonterminal symbol only expresses the order.

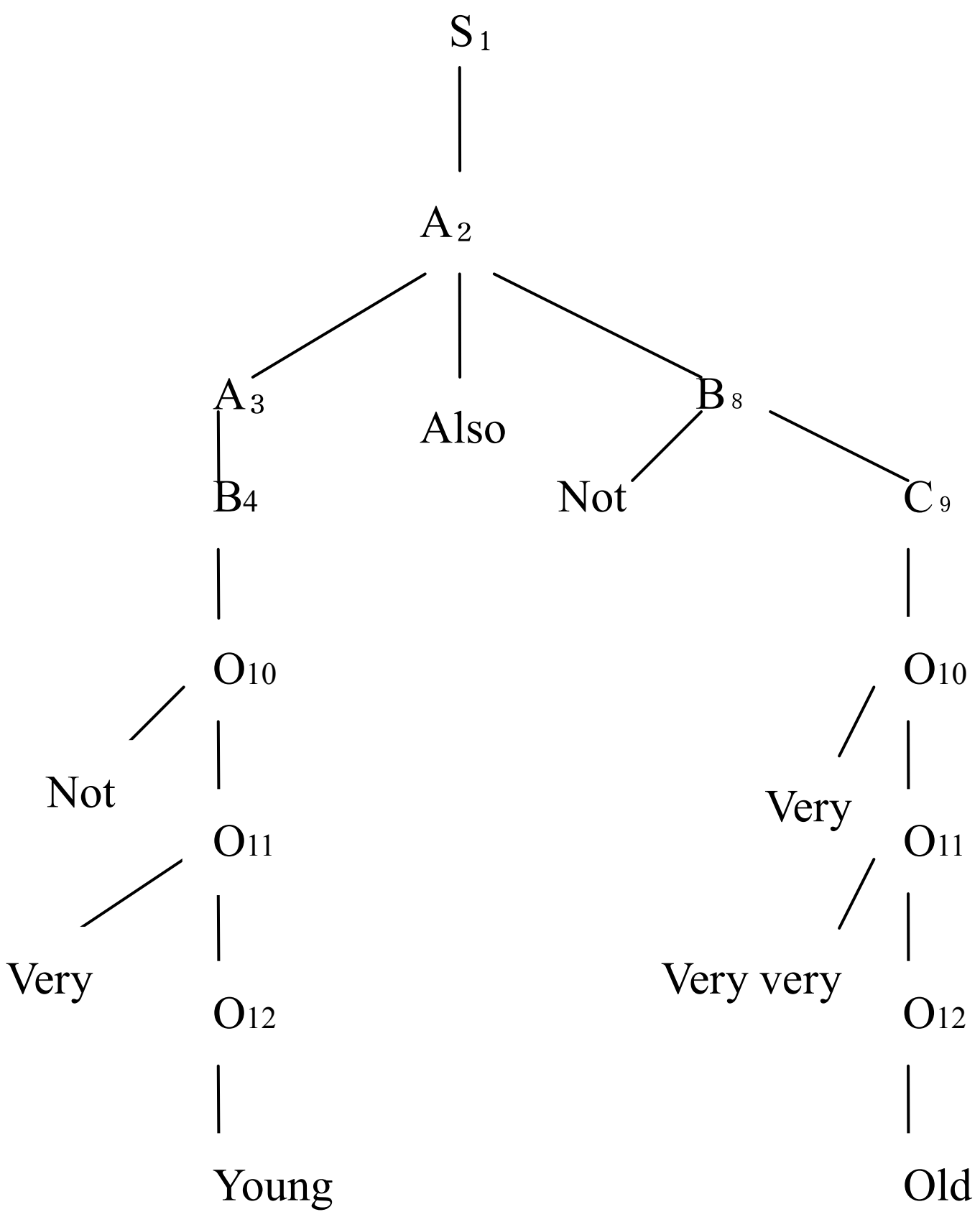

Figure 1.4.3 Grammar tree of 《neither quite young nor quite old》

The derivation of the membership function is as follows.

$$\mu(Y_7) = \mu_{young}(x)$$
$$\mu(Y_6) = \mu^2(Y_7)$$
$$\mu(C_5) = \mu(Y_6)$$
$$\mu(B_4) = 1 - \mu(C_5)$$
$$\mu(A_3) = \mu(B_4)$$
$$\mu(O_{12}) = \mu_{old}(x)$$
$$\mu(O_{11}) = \mu^2(O_{12})$$

$$\mu(O_{10}) = \mu^2(O_{11})$$
$$\mu(C_9) = \mu(O_{10})$$
$$\mu(B_8) = 1 - \mu(C_9)$$
$$\mu(A_2) = \mu(A_3) \wedge \mu(B_2)$$
$$\mu(S) = \mu(A_2)$$

Substituting these equations one by one, we can get the following result.

$$\mu(S) = (1 - \mu_{young}^2(x)) \wedge (1 - \mu_{old}^4(x))$$

# 1.5 Fuzzy Reasoning

## 1.5.1 General basis of reasoning

We often reason in our life, where the proposition containing the vague concept as the meaning of fuzziness is the premise.

For example,

Premise 1: If the tomato is red, it is ripe.

Premise 2: This tomato is very red.

Consequent: This tomato is well ripe.

Like this, it is very interesting to formulize the method that can do fuzzy reasoning mechanically in the view of natural language processing.

But, such reasoning is difficult by using the classical reasoning rule of binary logic. Therefore, new reasoning rule, compositional rule of reasoning was formulated and on the basis on it, fuzzy reasoning (approximate reasoning) was introduced.

Fuzzy reasoning is the reasoning deriving a certain new fuzzy proposition from several fuzzy propositions, and it is similar to the reasoning of human, so it can be applied to various branches such as fuzzy control, expert system, decision making and so on.

Reasoning is done by using the logic phenomenon such as 《If $P$ then $Q$》 $(P \to Q)$ (this is called implication.)

First, consider the logic operation $P \to Q$.

In case of binary logic,

| $P$ | $Q$ | $P \to Q$ |
|---|---|---|
| 1 | 1 | 1 |
| 1 | 0 | 0 |
| 0 | 1 | 1 |
| 0 | 0 | 1 |

$$P \to Q \triangleq \neg P \vee Q$$

where the truth value of $P, Q$ is denoted as $|P|, |Q|$, then $|P \to Q|$ is

$$|P \to Q| = (1 - |P|) \vee |Q| = |\neg P \text{or} Q| \tag{1.5.1}$$

In case of infinite value logic system,

$$|P - Q| = (1 - |P| + |Q|) \wedge 1 \tag{1.5.2}$$

At this time, using the boundary sum operation of fuzzy set $\oplus$, the following equation becomes as follows.

$$|P \to Q| = |\neg P| \oplus |Q|$$

From the (1-1) and (1-2), we can know that that $P \to Q$ is true means $|P| \le |Q|$.

Because if $|P \to Q| = r$, then (1.5.2) is written as follows.

$$r = (1 - |P| + |Q|) \wedge 1$$

And if $r = 1$, then

$$1 = (1 - |P| + |Q|) \wedge 1 \le (1 - |P| + |Q|)$$

, and finally $|P| \le |Q|$ is satisfied.

In the reasoning using $P \to Q$, there are 《modus ponens》 and 《modus tollens》.

① Modus Ponens

It is the method that reasons in case 《If *P* then *Q*》 is true, if P is true then Q is also true.

This is expressed as (1.5.3).

$$\begin{array}{l} 1^\circ\ P \to Q \\ 2^\circ\ P \\ \hline 3^\circ\ Q \end{array} \qquad (1.5.3)$$

In this equation, $P \to Q$ is called conditional sentence, *P* of $1^\circ$ is condition of conditional sentence, *Q* is conclusion, *P* of $2^\circ$ is antecedent, and *Q* of $3^\circ$ is consequent.

Modus ponens schema (1.5.3) means the following logic expression.

$$(P \text{ and } (P \to Q)) \to Q \qquad (1.5.4)$$

The validity of this reasoning is in that the truth value of logic expression (1.5.4) is always 1 regardless of the truth value of *P* and *Q*.

Verify the result of reasoning by modus ponens in case of binary logic.

By the assumption of the modus ponens, if $|P \to Q| = 1, |P| = 1$, then (1.5.4) becomes

$$1 = \neg(|P| \wedge |P| \to |Q|) \vee |Q| = 0 \vee |Q|$$

and therefore $|Q|$ must be 1.

② Modus tollens

Modus tollens is expressed as the following schema in the opposite form of modus ponens.

$$\begin{array}{l} 1^\circ\ P \to Q \\ \underline{2^\circ\ \neg Q} \\ 3^\circ \neg P \end{array} \qquad (1.5.5)$$

This schema reasons If 《If $P$, then $Q$》 is true and $\neg Q$ is true ($Q$ is false), then $\neg P$ is true ($P$ is false).

Schema (1.5.5) is expressed as the following logic expression.

$$(\neg Q \text{and} (P \to Q)) \to \neg P \qquad (1.5.6)$$

Reasoning in multi-value logic is done in the way of obtaining the truth value of $Q$, given the truth value of $P \to Q$ and $P$.

In the fuzzy logic system, language values are the truth value of propositions as mentioned before. And schema (1.5.3) and (1.5.5) are generalized as follows.

$$\begin{array}{l} 1^\circ\ P_1 \to Q_1 \\ \underline{2^\circ\ P_2 \qquad} \\ 3^\circ\ \ Q_2 \end{array}$$

$$\begin{array}{l} 1^\circ\ P_1 \to Q_1 \\ \underline{2^\circ\ \ Q_2\quad} \\ 3^\circ\ P_2 \end{array}$$

In these reasoning forms, the former is called fuzzy modus ponens, and the latter is fuzzy modus tollens.

The characteristics of these fuzzy reasoning are as follows.

First, in the conditional sentence, $P_1$ and $P_2$, $Q_1$ and $Q_2$ may be different.

Second, $P_1$, $P_2$, $Q_1$, $Q_2$ are expressed as fuzzy propositions, which are fuzzy sets.

## 1.5.2 IF-THEN fuzzy reasoning

The basic form of the general reasoning can be said that when the fuzzy relation *R* between a certain object *x* and *y* as knowledge and *A* that is the information of *x*, information of *y* from *R* and *A* is derives by reasoning.

Here, consider how to express the $P \to Q$ (If P then *Q*) as the fuzzy set in case *P* and *Q* are fuzzy propositions.

Consider the following two kinds of reasoning.

Fuzzy modus ponens

$$\begin{array}{l} 1^\circ\ \text{If } x \text{ is } A \text{ then } y \text{ is } B. \\ \underline{2^\circ\ x \text{ is } A'.\qquad\qquad\qquad} \\ \text{conclusion : } y \text{ is } B'. \end{array} \tag{1.5.7}$$

Modus tollens

$$\begin{array}{l} 1^\circ\ \text{If } x \text{ is } A \text{ then } y \text{ is } B. \\ 2^\circ\ y \text{ is } B'. \end{array} \tag{1.5.8}$$

---

conclusion：$x$ is $A'$.

where $x,y$ are the name of the objects and $A,\ A',\ B,\ B'$ are the fuzzy concepts, which are characterized by fuzzy sets in $U,\ V$.

Such fuzzy reasoning style is based on the idea converting the fuzzy conditional sentence of $1^\circ$ 《If $x$ is $A$, then $y$ is $B$》 to the fuzzy relation in fuzzy modus ponens and fuzzy modus tollens.

That is,

$$《x \text{ is } A》 \rightarrow 《y \text{ is } B》 = 《(x,y) \text{ is } R》. \qquad (1.5.9)$$

By expressing like this, we can consider $(x,y)$ to the names of objects, R is the predicate, and this is a proposition.

Referring $A,\ B$ to the fuzzy sets of $U,\ V$ respectively, and denoting 《If $x$ is $A$, then $y$ is $B$.》 as $A \rightarrow B$, $R = (A \rightarrow B)$ is defined as follows.

$$\mu_{A\rightarrow B}(u,\ v) = \mu_A(u) \rightarrow \mu_B(v)$$

At this time, $R$ may be changed according to what implication $A \rightarrow B$ is used as.

As typical example,

$$R_c = A \times B = \int_{U\times V} \mu_A(u) \wedge \mu_B(v)/(u,v) \qquad (1.5.10)$$

$$R_m = (A \times B) \bigcup (\neg A \times V) = \int_{U\times V} [(\mu_A(u) \wedge \mu_B(v)) \wedge (1 - \mu_A(u))]/(u,v) \qquad (1.5.11)$$

$$R_a = \neg A \oplus B = \int_{U\times V} [1 \wedge (1 - \mu_A(u) + \mu_B(v))]/(u,v) \qquad (1.5.12)$$

where $R_a$ is by $|P \rightarrow Q| = |\neg P| \oplus |Q|$. $R_a$ represents the truth value of $P \rightarrow Q$,

but $R_a$ means $A \to B$.

$R_c$ may be not the expression of $\to$, but it is much used in fuzzy control.

Expression of $R_m$ has its logical meaning. It is as follows.

Antecedent proposition 《If *x* is *A*, then *y* is *B*》 has no its meaning by itself logically. That is, adding a certain condition is for distinction between the other conditions, and furthermore besides them, results express the others.

Therefore, accommodation 《If *x* is *A*, then *y* is *B*》 has no meaning by combining the $\neg$ with the other condition.

So, consider the accommodation 《If *x* is $\neg A$, then *y* is no》.

《is no》 is expresses as the whole set *V* which element is *y*. then 《*x* is *A*→ *y* is *B*》 is analyzed as follows.

《$x$ is $A \to y$ is B》 or 《$x$ is $\neg A \to y$ is V》 = 《$(x,y)$ is $(A \to B) \cup (\neg A \to V)$》

We can understand that using $R_c$ as $A \to B$, fuzzy relation of the right of the above equation is $R_m$.

In fuzzy modus ponens, there are two kinds, direct method and indirect method, consider only direct method.

Direct method is the method reasoning the $B'$ directly from $A \to B$.

Fuzzy reasoning of direct method consists of compositional rule of fuzzy relation. That is,

$$B' = (A \to B) \circ A' \tag{1.5.13}$$

This is called compositional rule of fuzzy reasoning. Expressing this compositional rule with the membership function is as follows.

$$\mu_{B'}(v) = \vee_u \{\mu_{A'}(u) \wedge \mu_{A\to B}(u,v)\} = \vee_u \{\mu_{A'}(u) \wedge [\mu_A(u) \to \mu_B(v)]\} \quad (1.5.14)$$

Using fuzzy relation $R_c$, $R_m$, $R_a$ instead of $(A \to B)$, $B'$ is obtained as follows.

$$B'_c = A' \circ R_c = A' \circ (A \times B) \quad (1.5.15)$$

$$B'_m = A' \circ R_m = A' \circ [(A \times B) \cup (\neg A \times V)] \quad (1.5.16)$$

$$B'_a = A' \circ R_a = A' \circ (\neg A \oplus B) = A' \circ [(\neg A \times V) \oplus (U \times B)] \quad (1.5.17)$$

In the same way, conclusion by fuzzy modus tollens $A'$ is obtained as follows for $B'$.

$$A'_c = R_c \circ B' = (A \times B) \circ B' \quad (1.5.18)$$

$$A'_m = R_m \circ B' = [(A \times B) \cup (\neg A \times V)] \circ B' \quad (1.5.19)$$

$$A'_a = R_a \circ B' = [(\neg A \times v) \oplus (U \times B)] \circ B' \quad (1.5.20)$$

Consider $R_c$ for example. That is, this is the case $| a \to b | = 0 \wedge b$.

But note that this doesn't satisfy the property of implication because

$$| 0 \to b | = 0 \wedge b = 0.$$

From

$B'_c = A' \circ R_c = A' \circ (A \times B)$,

$$\mu_{B'}(v) = \vee_u \{\mu_{A'}(u) \wedge [\mu_A(u) \wedge \mu_B(v)]\} = \vee_u \{\mu_{A'}(u) \wedge \mu_A(u)\} \wedge \mu_B(v) = h \wedge \mu_B(v)$$

This is shown in figure 1.5.1.

Here h is the maximum height of the product of fuzzy sets $A'$ and $A$, and shows how much $A'$ and $A$ are related. As result, $B'$ is obtained through the minimum operation of $h$ and $B$. that is, the upper part of $B$ is cut by $h$ is $B'$.

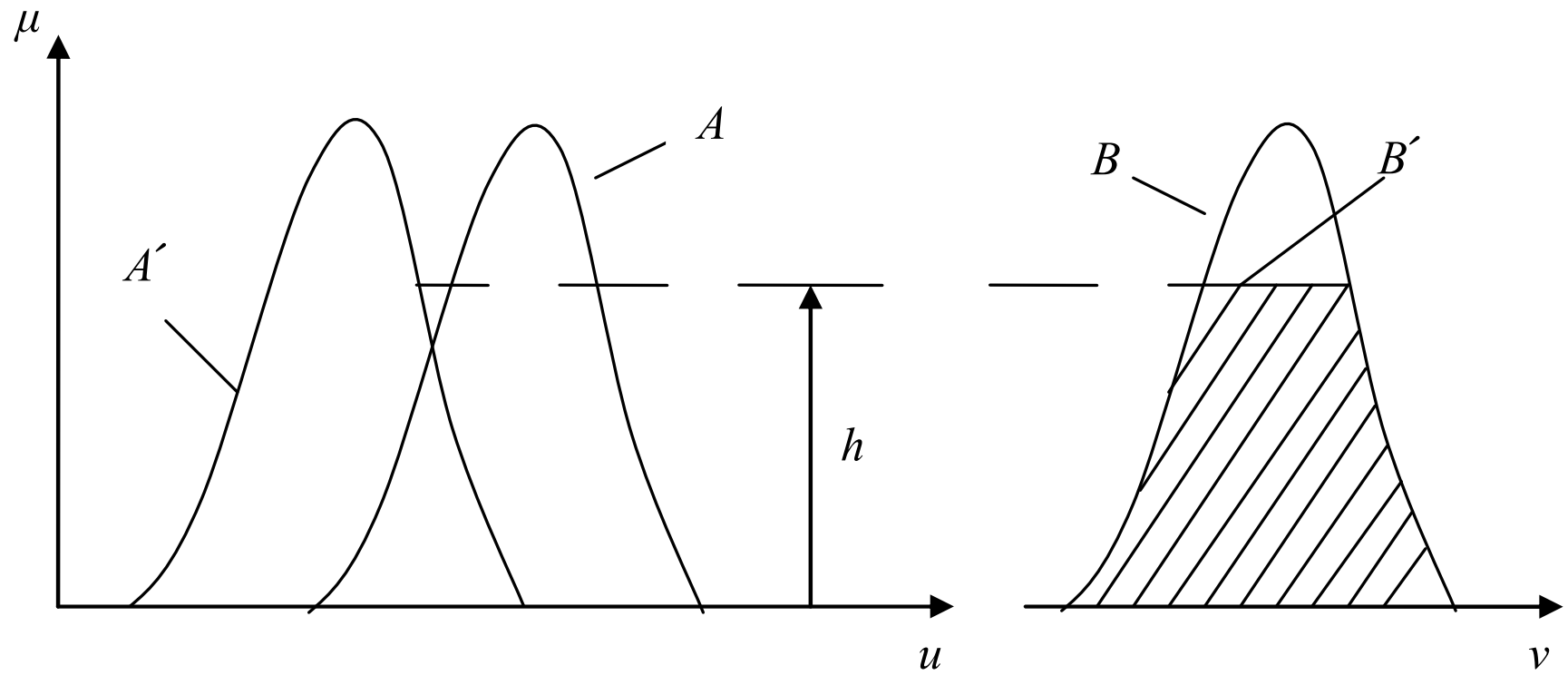

Figure 1.5.1 Reasoning result from $A'$ and $A \to B$

Like this, fuzzy reasoning derives the proper result not only in case that $A'$ and $A$ are completely consistent and not but also according to the degree of consistency of $A'$ and $A$.

If $A'$ and $A$ are completely consistent, $h = 1$. So consequent $B'$ is consistent with $B$, and satisfies the modus ponens.

In fuzzy control theory, in most cases, $A'$ is the finite value (for example, bias).

That is, in case $A' = u_0.$, $\mu_A(u_0) = 1, \mu_A(u) = 0 (u \neq u_0)$ is possible, and the consequent $B'$ is as follows.

$$\mu_{B'}(v) = \underset{u}{\vee}\{\mu_{A'}(u) \wedge [\mu_A(u) \to \mu_B(v)]\} = \underset{u \neq u_1}{\vee}\{0 \wedge [\mu_A(u) \to \mu_B(v)]\} \vee$$

$$\vee\{1 \wedge [\mu_A(u_0) \to \mu_B(v)]\} = \mu_A(u_0) \to \mu_B(v)$$

Therefore,

$$\mu_{B'}(v) = \mu_A(u_0) \to \mu_B(v)$$

This means $h = \mu_A(u_0)$.

There are $R_g, R_s$ and their hybrid methods beside fuzzy relation $R_c, R_m, R_a$ in expressing the fuzzy relation $P \to Q$.

Referring $A,B$ to the fuzzy sets of $U,\ V$ respectively,

$$R_s = A\times V \xrightarrow[S]{} U\times B = \int_{U\times V} [\mu_A(u) \xrightarrow[S]{} \mu_B(v)]/(u,v) \qquad (1.5.21)$$

where

$$\mu_A(u) \xrightarrow[S]{} \mu_B(v) = \begin{cases} 1: & \mu_A(u) \le \mu_B(v) \\ 0: & \mu_A(u) > \mu_B(v) \end{cases} \qquad (1.5.22)$$

$$R_g = A\times V \xrightarrow[g]{} U\times B = \int_{U\times V} [\mu_A(u) \xrightarrow[g]{} \mu_B(v)]/(u,v) \qquad (1.5.23)$$

where

$$\mu_A(u) \xrightarrow[g]{} \mu_B(v) = \begin{cases} 1: & \mu_A(u) \le \mu_B(v) \\ \mu_B(v): & \mu_A(u) > \mu_B(v) \end{cases} \qquad (1.5.24)$$

By combining $R_s$ and $R_g$, the following fuzzy relation is obtained.

$$R_{sg} = (A\times V \xrightarrow[s]{} U\times B) \bigcap (\neg A\times V \xrightarrow[g]{} U\times \neg B) =$$

$$= \int_{U\times V} (\mu_A(u) \xrightarrow[s]{} \mu_B(v)) \wedge ((1-\mu_A(u)) \xrightarrow[g]{} (1-\mu_B(v)))/(u,\ v) \qquad (1.5.25)$$

$$R_{gg} = (A\times V \xrightarrow[g]{} V\times B) \bigcap (\neg A\times V \xrightarrow[g]{} U\times \neg B) =$$

$$= \int_{U\times V} (\mu_A(u) \xrightarrow[g]{} \mu_B(v)) \wedge ((1-\mu_A(u)) \xrightarrow[g]{} (1-\mu_B(v)))/(u,\ v) \qquad (1.5.26)$$

$$R_{gs} = (A\times V \xrightarrow[g]{} U\times B) \bigcap (\neg A\times V \xrightarrow[s]{} U\times \neg B) =$$

$$= \int_{U\times V} (\mu_A(u) \xrightarrow[g]{} \mu_B(v)) \wedge ((1-\mu_A(u)) \xrightarrow[s]{} (1-\mu_B(v)))/(u,\ v) \qquad (1.5.27)$$

$$R_{ss} = (A\times V \xrightarrow[s]{} U\times B) \bigcap (\neg A\times V \xrightarrow[s]{} U\times \neg B) =$$

$$= \int_{U\times V} (\mu_A(u) \xrightarrow[s]{} \mu_B(v)) \wedge ((1-\mu_A(u)) \xrightarrow[s]{} (1-\mu_B(v)))/(u,\ v) \qquad (1.5.28)$$

And by introducing the accommodation in multi-value logic, the following fuzzy relation is obtained.

$$R_{\#} = (\neg A \times V) \bigcup (U \times B) = \int_{U \times V} [(1 - \mu_A(u)) \vee \mu_B(v)]/(u,\ v) \tag{1.5.29}$$

$$R_{\Delta} = A \times V \xrightarrow[\Delta]{} U \times B = \int_{U \times V} [\mu_A(u) \xrightarrow[\Delta]{} \mu_B(v)]/(u,\ v) \tag{1.5.30}$$

where

$$\mu_A(u) \xrightarrow[\Delta]{} \mu_B(v) = \begin{cases} 1; & \mu_A(u) \le \mu_B(v) \\ \dfrac{\mu_B(v)}{\mu_A(u)}; & \mu_A(u) > \mu_B(v) \end{cases} \tag{1.5.31}$$

$$R_{\square} = A \times V \xrightarrow[\square]{} U \times B = \int_{U \times V} [\mu_A(u) \xrightarrow[\square]{} \mu_B(v)]/(u,\ v) \tag{1.5.32}$$

where

$$\mu_A(u) \xrightarrow[\square]{} \mu_B(v) = \begin{cases} 1: \ \mu_A(u) \ne 1, \ or \ \mu_B(v) = 1, \\ 0: \ \text{others} \end{cases} \tag{1.5.33}$$

$$R_{*} = A \times V \xrightarrow[*]{} U \times B = \int_{U \times V} 0 \vee (\mu_B(v) \to \mu_A(u))/(u,\ v). \tag{1.5.34}$$

$$\mu_A(u) \xrightarrow[*]{} \mu_B(v) = 0 \vee (\mu_A(u) \wedge \mu_B(v)) \tag{1.5.35}$$

Among the 13 fuzzy relations defined above, $R_m, R_a, R_c$ are shown in figure 1.5.2, 1.5.3, 1.5.4.

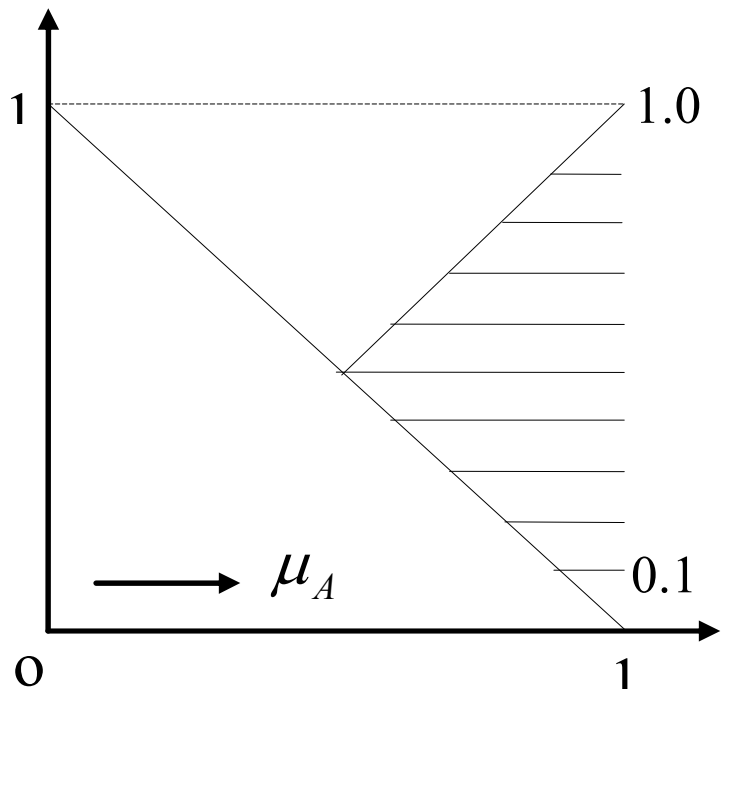

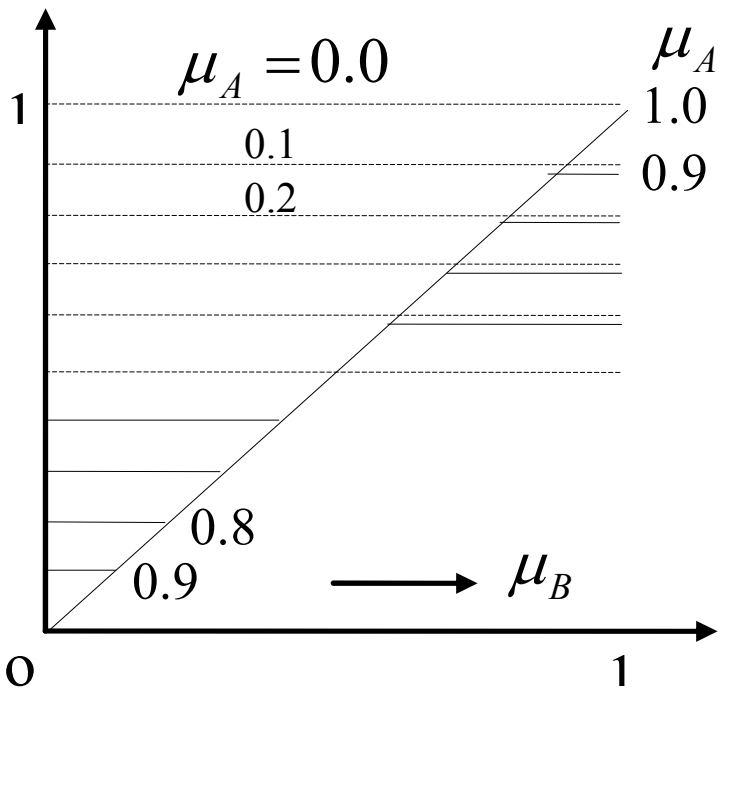

Figure 1.5.2 Fuzzy relation $R_m$

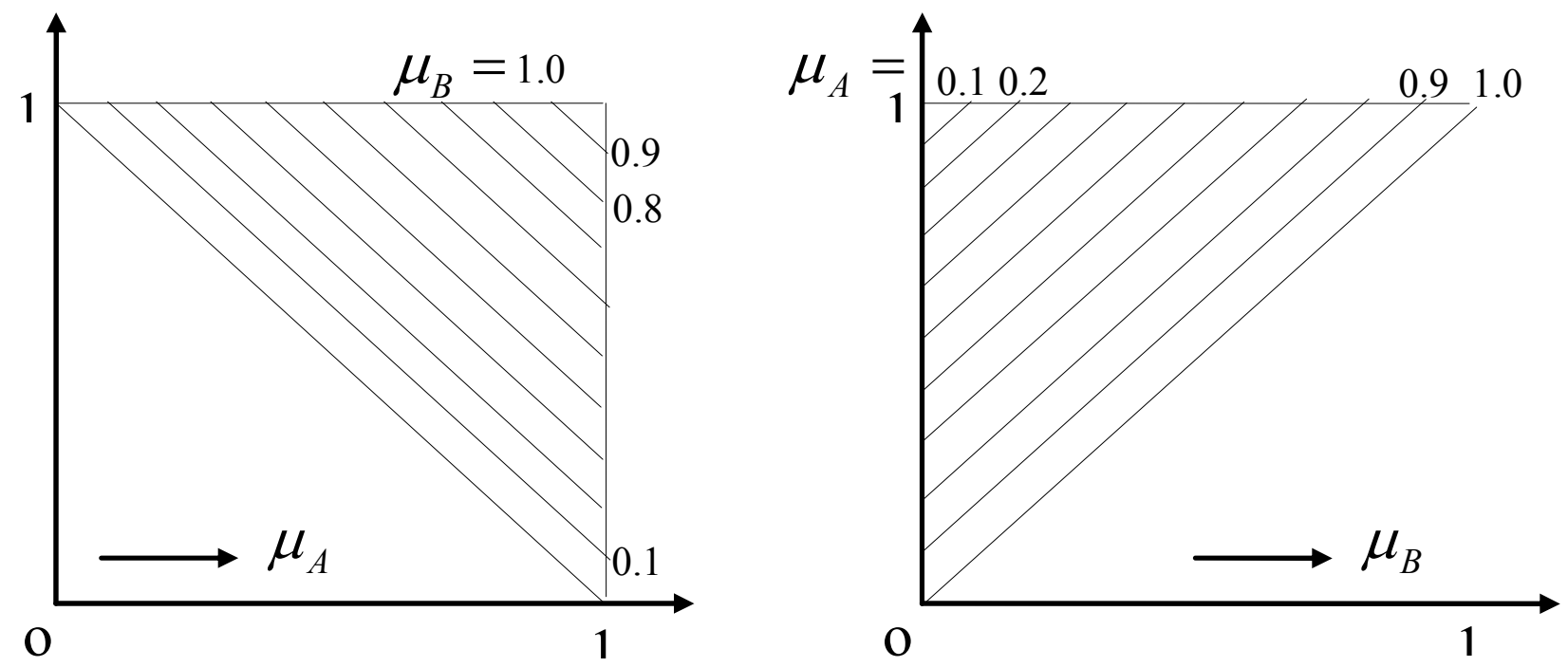

Figure 1.5.3 Fuzzy relation $R_a$

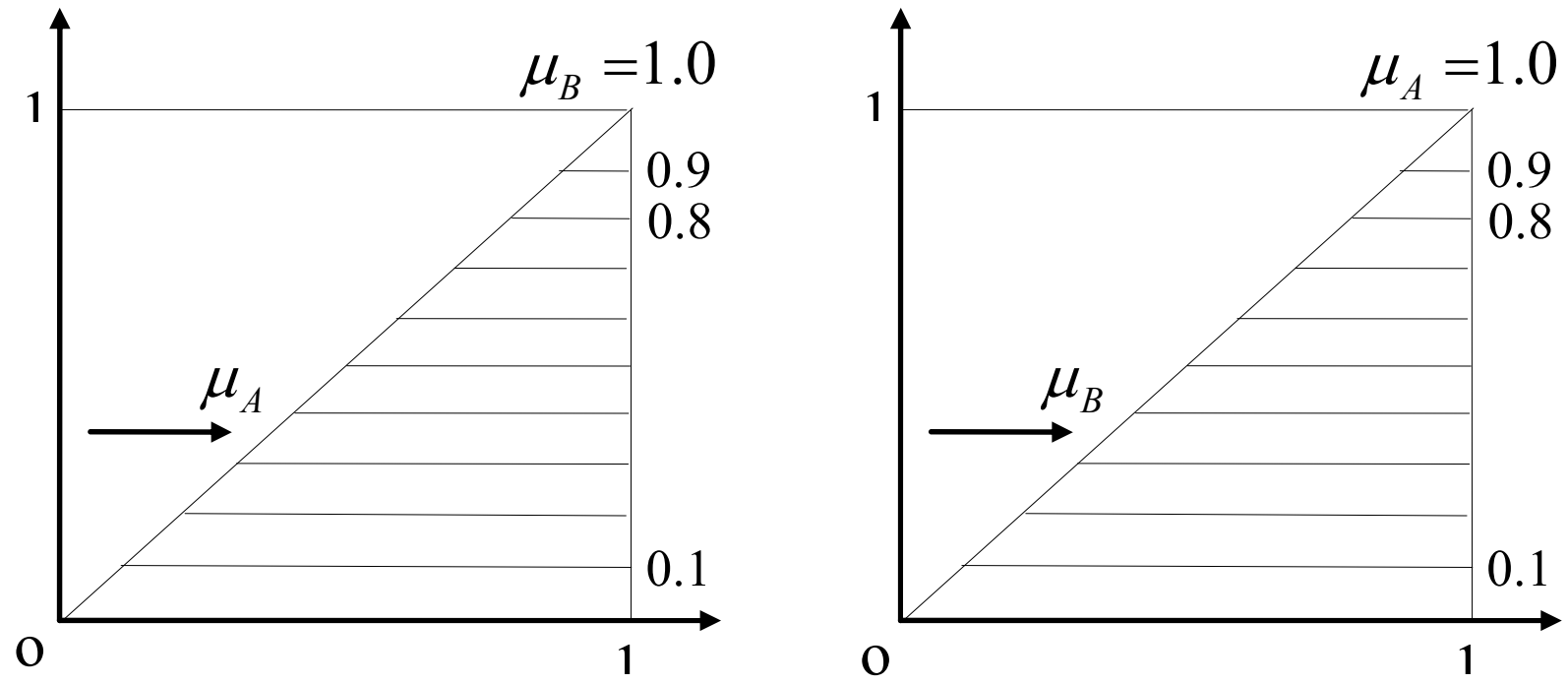

Figure 1.5.4 Fuzzy relation $R_c$

In each figure, figure drawn with $\mu_B(v)$ as parameter is corresponding to the fuzzy modus ponens, and figure with $\mu_A(u)$ is to fuzzy modus tollens. And $\mu_A, \mu_B$ mean $\mu_A(u),\ \mu_B(v)$.

Apply the above fuzzy relations to fuzzy modus ponens and fuzzy modus tollens. In fuzzy modus ponens of (1.5.6), referring $A'$ in 2° to $A$, very $A(=A^2)$, more or less $A(=A^{0.5})$, not $A(=\neg A)$, then calculate the consequent $B'$ using the fuzzy relation and composition operator.

In the same way, in fuzzy modus tollens in (1.5.8), referring $B'$ in $2^\circ$ to not $B$, not very $B$, not more or less $B$ respectively, calculate consequent $A'$. The results are shown in table 1.5.1.

Table 1.5.1. Fuzzy reasoning result.

(a) Fuzzy modus ponens

| Relations | $A$ | very $A$ | more or less $A$ | not $A$ | Reductive property, % |
|---|---|---|---|---|---|
| $R_m$ | $0.5 \vee \mu_B$ | $\frac{1}{2}(\sqrt{5}-1) \vee \mu_B$ | $\frac{1}{2}(3-\sqrt{5}) \vee \mu_B$ | 1 | 0 |
| $R_a$ | $\frac{1}{2}(1+\mu_B)$ | $\frac{1}{2}(3+2\mu_B - -\sqrt{5-4\mu_B})$ | $\frac{1}{2}\sqrt{5+4\mu_B-1}$ | 1 | 0 |
| $R_c$ | $\mu_B$ | $\mu_B$ | $\mu_B$ | $0.5 \wedge \mu_B$ | 25 |
| $R_s$ | $\mu_B$ | $\mu_B^2$ | $\sqrt{\mu_B}$ | 1 | 75 |
| $R_g$ | $\mu_B$ | $\mu_B$ | $\sqrt{\mu_B}$ | 1 | 50 |
| $R_{sg}$ | $\mu_B$ | $\mu_B^2$ | $\sqrt{\mu_B}$ | $1-\mu_B$ | 100 |
| $R_{gg}$ | $\mu_B$ | $\mu_B$ | $\sqrt{\mu_B}$ | $1-\mu_B$ | 75 |
| $R_{gs}$ | $\mu_B$ | $\mu_B$ | $\sqrt{\mu_B}$ | $1-\mu_B$ | 75 |
| $R_{ss}$ | $\mu_B$ | $\mu_B^2$ | $\sqrt{\mu_B}$ | $1-\mu_B$ | 100 |
| $R_\#$ | $0.5 \vee \mu_B$ | $\frac{1}{2}(3-\sqrt{5}) \vee \mu_B$ | $\frac{1}{2}(\sqrt{5}-1) \vee \mu_B$ | 1 | 0 |
| $R_{\Delta}$ | $\mu_{\mathrm{B}}^{0.5}$ | $\mu_B^2$ | $\sqrt[3]{\mu_B}$ | 1 | 25 |
| $R_{\square}$ | 1 | 1 | 1 | 1 | 0 |
| $R_{*}$ | $\frac{1}{2}\mu_B$ | $\frac{1}{2}(2\mu_B+1-$ $-\sqrt{4\mu_B+1})$ | $\frac{1}{2}(\sqrt{1+4\mu_B}-1)$ | $\mu_B$ | 10 |

(b) Fuzzy Modus tollens

| Relations | not $B$ | not very $B$ | not more or less $A$ | $B$ | Reductive property, % |
|---|---|---|---|---|---|
| $R_m$ | $0.5\vee\overline{\mu_A}$ | $\left(\overline{\mu_A}\vee\frac{\sqrt{5}-1}{2}\right)\vee\mu_A$ | $\frac{1}{2}(3-\sqrt{5})\vee$ $\vee(1-\mu_A)$ | $\mu_A\vee\overline{\mu}_A$ | 25 |
| $R_a$ | $1-\frac{\mu_A}{2}$ | $\frac{1}{2}(1-2\mu_A+$ $+\sqrt{1+4\mu_A})$ | $\frac{1}{2}(3-\sqrt{1+4\mu_A})$ | 1 | 0 |
| $R_c$ | $0.5\wedge\mu_A$ | $\frac{1}{2}(\sqrt{5}-1)\wedge\mu_A$ | $\frac{1}{2}(3-\sqrt{5})\wedge\mu_A$ | $\mu_A$ | 37.5 |
| $R_s$ | $1-\mu_A$ | $1-\mu_A^2$ | $1-\mu_A^{0.5}$ | 1 | 75 |
| $R_g$ | $0.5\vee\overline{\mu_A}$ | $\frac{\sqrt{5}-1}{2}\vee(1-\mu_A^2)$ | $\frac{3-\sqrt{5}}{2}\vee(1-\mu_A^{0.5})$ | 1 | 62.5 |
| $R_{sg}$ | $1-\mu_A$ | $1-\mu_A^2$ | $1-\mu_A^{0.5}$ | $0.5\vee\mu_A$ | 87.5 |
| $R_{gg}$ | $0.5\vee\overline{\mu_A}$ | $\frac{\sqrt{5}-1}{2}\vee(1-\mu_A^2)$ | $\frac{3-\sqrt{5}}{2}\vee(1-\mu_A^{0.5})$ | $0.5\vee\mu_A$ | 50 |
| $R_{gs}$ | $0.5\vee\overline{\mu_A}$ | $\frac{\sqrt{5}-1}{2}\vee(1-\mu_A^2)$ | $\frac{\sqrt{5}-1}{2}\vee(1-\mu_A^2)$ | $\mu_A$ | 50 |
| $R_{ss}$ | $1-\mu_A$ | $1-\mu_A^2$ | $1-\mu_A^{0.5}$ | $\mu_A$ | 100 |
| $R_{\#}$ | $0.5\vee\overline{\mu_A}$ | $\frac{\sqrt{5}-1}{2}\vee(1-\mu_A^2)$ | $\frac{3-\sqrt{5}}{2}\vee(1-\mu_A^{0.5})$ | 1 | 12.5 |
| $R_{\Delta}$ | $\frac{1}{1+\mu_A}$ | $\frac{\sqrt{1+\mu_A^2}-1}{2\mu_A^2}$ | $\frac{1+\mu_A-\sqrt{\mu_A^2+4\mu_A}}{2}$ | 1 | 0 |
| $R_{\square}$ | $\begin{cases}1, & \mu_A<1\\0, & \mu_A=0\end{cases}$ | $\begin{cases}1, & \mu_A<1\\0, & \mu_A=0\end{cases}$ | $\begin{cases}1, & \mu_A<1\\0, & \mu_A=0\end{cases}$ | 1 | 0 |
| $R_*$ | $\frac{1-\mu_A}{2}$ | $\frac{\sqrt{5+4\mu_A-(2+\mu_A)}}{2}$ | $\frac{3-\sqrt{5+4\mu_A}}{2}$ | $1-\mu_A$ | 12.5 |

(Example 1.5.1) For example, consider Fuzzy relation $R_\#$.

By the composition of Fuzzy relation $R_\#$ and $A'$, the following equation is satisfied.

$$B'_\# = A \circ R_\# = A' \circ [(\neg A \times V) \cup (U \times B)] = \int_U (\mu_{A'}(u)/u) \circ \left\{ \int_{U\times V} [(1-\mu_A(u)) \vee \mu_B(v)]/(u,v) \right\} =$$

$$= \int_V \vee_u \{\mu_{A'}(u) \wedge [(1-\mu_A(u)) \vee \mu_B(v)]\}/v \qquad (1.5.36)$$

If $A'$ is very $A(=A^2)$, then $\mu_{A'} = \mu_A^2$ and $\mu_{B'_\#}$ is shown in figure 1.5.5.

From this, $\mu_{A'} \wedge [(1-\mu_A) \vee \mu_B]$ in (1.5.36) is the same as the dotted line of figure 1.5.8 in case $\mu_B = 0.2$, and the maximum value is $(3-\sqrt{5})/2$.

In the same way, in case $\mu_B = 0.7$, maximum value is o.7.

After all, $\mu_{B'_\#}$ is

$$\mu_{B'_\#} = \begin{cases} \frac{3-\sqrt{5}}{2}, \mu_B \le \frac{3-\sqrt{5}}{2}, \\ \mu_B, \mu_B > \frac{3-\sqrt{5}}{2} \end{cases}$$

or $\mu_{B'_\#} = \frac{3-\sqrt{5}}{2} \vee \mu_B$, and then fuzzy set of final reasoning result.

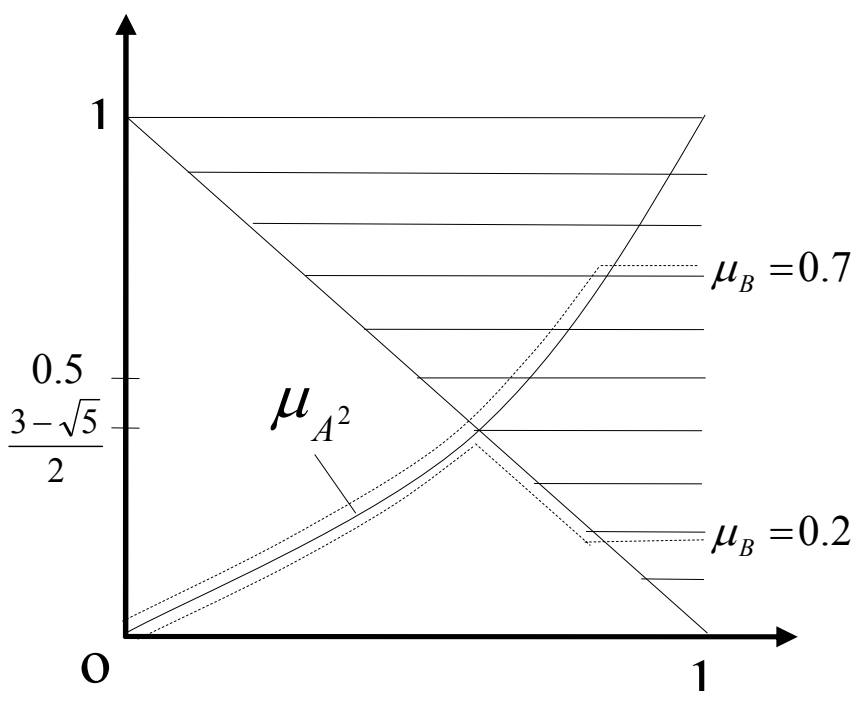

Figure 1.5.5 Method of obtaining $\mu_{B'_\#}$ ( in case $\mu_{A'} = \mu_{A^2}$ )

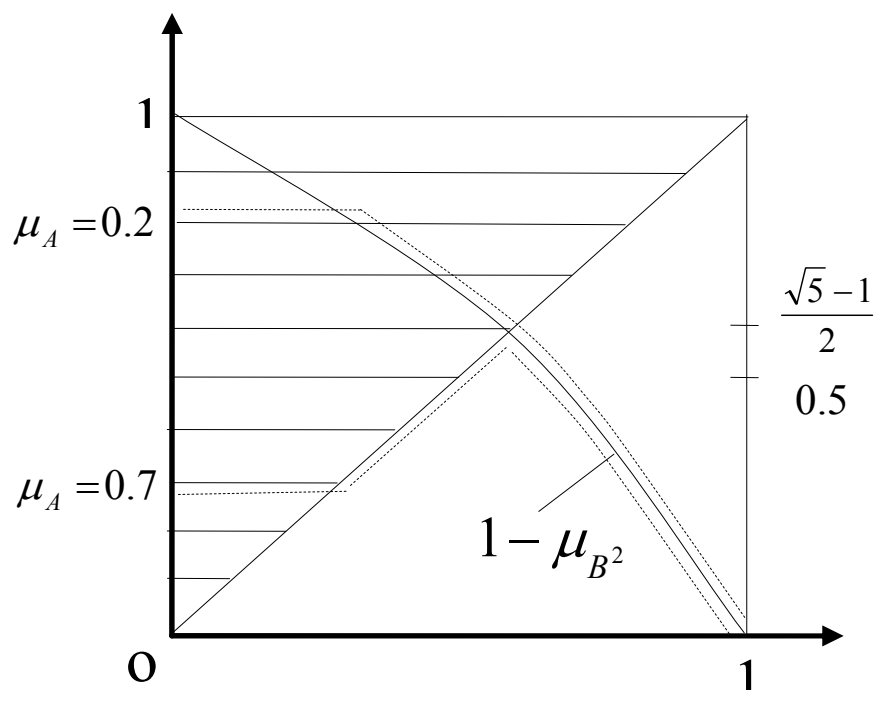

Figure 1.5.6 Method obtaining $\mu_{A'_\#}$ ( in case $\mu_{B'} = 1-\mu_{B^2}$ )

In the same way, we can consider the case of $\mu_{A'} = \mu_A, \mu_A^{0.5}, (1-\mu_A)$.

Next, consider the case of fuzzy modus tollens.

$$A_{\#} = R_\# \circ B' = [(\neg A \times V) \cup$$

$$\cup (U \times B)] \circ B' =$$

$$= \int_U \vee_v [(1-\mu_A) \vee \mu_B \wedge \mu_{B'}]/u$$

If $B' =$ not Very $B(= \neg B^2)$, then $\mu_{B'} = 1 - \mu_B^2$ and $\mu_{A'_\#}$ is the same as figure 1.5.6.

In this figure, $A'_\#$ is the same as the dotted line in case $\mu_A = 0.2$, and the maximum value is 0.8.

In case, $\mu_A = 0.7$, it is $\frac{\sqrt{5}-1}{2}$.

Therefore,

$$\mu_{A'_\#} = \begin{cases} \frac{\sqrt{5}-1}{2}, & \mu_A \geq \frac{\sqrt{5}-1}{2} \\ 1-\mu_A, & \mu_A < \frac{\sqrt{5}-1}{2} \end{cases}$$

and the following equation is satisfied.

$$\mu_{A'_\#} = \frac{\sqrt{5}-1}{2} \vee (1-\mu_A).$$

In the same way, we can consider the case of $\mu_{B'} = 1-\mu_B,\ 1-\mu_B^{0.5},\ \mu_B$.

### 1.5.3 Comparison between syllogism and fuzzy reasoning

Then, consider whether syllogism is satisfied when we use the various fuzzy relation considered above.

Syllogism, the generalization of the reasoning schema of (1.5.7) is formulated as follows in case of fuzzy conditional sentence $P_1,\ P_2, P_3$.

$$\begin{array}{l} P_1\text{: If } x \text{ is } A\text{, then } y \text{ is } B. \\ P_2\text{: If } y \text{ is } B\text{, then } z \text{ is } C. \\ \hline P_3\text{: If } x \text{ is } A\text{, then } z \text{ is } C. \end{array} \qquad (1.5.37)$$

This is

$$\begin{array}{l} P_1\text{: } A \to B \\ P_2\text{: } B \to C \\ \hline P_3\text{: } A \to C \end{array} \qquad (1.5.38)$$

where $A,B,C$ are the fuzzy sets of $U,V,W$ .

If $A \to C$ is obtained as the result of composition of $A \to B$ and $B \to C$, that is

$$(A \to B) \circ (B \to C) = (A \to C) \tag{1.5.39}$$

, then syllogism is said to be satisfied.

Converting $A \to B,\ B \to C,\ A \to C$ to fuzzy relation $R_m$,

$$R_m(A,\ B) = (A \times B) \cup (\neg A \times V) \tag{1.5.40}$$

$$R_m(B,\ C) = (B \times C) \cup (\neg B \times W) \tag{1.5.41}$$

$$R_m(A,\ C) = (A \times C) \cup (\neg A \times W) \tag{1.5.42}$$

If syllogism is held, then

$$R_m(A,B) \circ R_m(B,C) = R_m(A,C) \tag{1.5.43}$$

The result applying this to the above 13 fuzzy relation is as follows.

| Fuzzy Relations | $R_m$ | $R_a$ | $R_c$ | $R_s$ | $R_g$ | $R_{sg}$ | $R_{gg}$ | $R_{gs}$ | $R_{ss}$ | $R_\#$ | $R_\Delta$ | $R_\square$ | $R_*$ |
|---|---|---|---|---|---|---|---|---|---|---|---|---|---|
| Syllogism | × | × | × | O | O | O | O | O | O | × | × | × | × |

where O expresses that fuzzy relation satisfies the syllogism, and × expresses that it doesn’t .

## 1.5.4 IF-THEN-ELSE type fuzzy reasoning

We consider the fuzzy reasoning such as 《If $x$ is $A$ then $y$ is $B$ else $y$ is $C$》 as the generalization of fuzzy reasoning, and this has the premise 《If $x$ is $A$ then $y$ is B》 .

For example of the generalization, 《If demand is great, then the cost is expensive, else it is inexpensive.》

Diagramming such a reasoning type, the result is as follows.

$$\begin{array}{ll} 1^\circ \ \text{if } x \text{ is } A \text{ then } y \text{ is } B \text{ else } y \text{ is } C & \\ 2^\circ \ x \text{ is } A' & (1.5.44) \\ \text{conclusion } y \text{ is } D & \end{array}$$

where $x$, $y$ are the names of objects, $A$, $A'$ are the fuzzy sets in $U$, $B$, $C$, $D$ are the fuzzy sets in $V$.

Considering the fuzzy conditional sentence of $1^\circ$ 《If $x$ is $A$ then $y$ is $B$ else $y$ is $C$》 in the above reasoning type, we can understand that this expresses the fuzzy relation between $A$ and $B,C$.

Fuzzy conditional sentence 《If $x$ is $A$ then $y$ is $B$ else $y$ is $C$》 can be written as 《(If $x$ is $A$ then $y$ is $B$) and (If $x$ is not $A$ then $y$ is $C$)》.

That is,

$$A \to B \text{ and } \neg A \to C.$$

From this, we can apply by composing the reasoning method of $A \to B$ as the case of fuzzy reasoning (If ~ then ~).

In reasoning type (1.5.10), there are several fuzzy relations representing $1^\circ$. Typically, $R'_m$, $R'_a$, $R'_b$ are defined as follows.

$$R'_m = (A \times B) \bigcup (\neg A \times C) = \int_{U \times V} (\mu_A(u) \wedge \mu_B(v)) \vee ((1-\mu_A(u)) \wedge \mu_c(v)) / (u,\ v)$$

$$R'_a = (\neg A \times V \oplus U \times B) \bigcap (A \times V \oplus U \times C) =$$

$$= \int_{V \times U} 1 \wedge (1-\mu_A(u)+\mu_B(v)) \wedge (\mu_A(u)+\mu_c(v)) / (u,\ v)$$

$$R'_b = (\neg A \times V \bigcup U \times B) \bigcap (A \times V \bigcup U \times C) =$$

$$= \int_{U\times V} ((1-\mu_A(u))\vee \mu_B(v))\wedge(\mu_A(u)\vee \mu_c(v))/(u,\ v) \tag{1.5.45}$$

Using these relations, conclusion $D$ of (1.5.44) is obtained by $A'$ and composition operator. Given $A'$ and $R'_m$,

$$D_m = A'\circ R'_m = A'\circ[(A\times B)\bigcup(\neg A\times C)] \tag{1.5.46}$$

And the membership function is as follows.

$$\mu_{D_m}(v) = \vee_u\{\mu_{A'}(u)\wedge[(\mu_A(u)\wedge \mu_B(v))\vee(1-\mu_A(u))\wedge \mu_c(v)]\} \tag{1.5.47}$$

In the same way,

$$D_a = A'\circ[(\neg A\times V\oplus U\times B)\bigcap(A\times V\oplus U\times C)] \tag{1.5.48}$$

$$D_b = A'\circ[(\neg A\times V\bigcup U\times B)\bigcap(A\times V\bigcup U\times C)] \tag{1.5.49}$$

And also, using the $R_g, R_s$ considered in 《If ~ then ~》 type, the following fuzzy relation can be obtained.

$$R'_{gg} = (A\times V \xrightarrow[g]{} U\times B)\bigcap(\neg A\times V \rightarrow U\times C) =$$

$$= \int_{U\times V} (\mu_A(u) \xrightarrow[g]{} \mu_B(v))\wedge((1-\mu_A(u)) \xrightarrow[g]{} \mu_c(v))/(u,\ v) \tag{1.5.50}$$

$$R'_{gs} = (A\times V \xrightarrow[g]{} U\times B)\bigcap(\neg A\times V \xrightarrow[s]{} U\times C) =$$

$$= \int_{U\times V} (\mu_A(u) \xrightarrow[g]{} \mu_B(v))\wedge((1-\mu_A(u)) \xrightarrow[s]{} \mu_c(v))/(u,\ v) \tag{1.5.51}$$

$$R'_{sg} = (A\times V \xrightarrow[s]{} U\times B)\bigcap(\neg A\times V \xrightarrow[g]{} U\times C) =$$

$$= \int_{U\times V} (\mu_A(u) \xrightarrow[s]{} \mu_B(v))\wedge((1-\mu_A(u)) \xrightarrow[g]{} \mu_c(v))/(u,\ v) \tag{1.5.52}$$

$$R'_{ss} = (A\times V \xrightarrow[s]{} U\times B)\bigcap(\neg A\times V \xrightarrow[s]{} U\times C) =$$

$$= \int_{U\times V} (\mu_A(u) \xrightarrow[s]{} \mu_B(v))\wedge((1-\mu_A(u)) \xrightarrow[s]{} \mu_c(v))/(u,\ v) \tag{1.5.53}$$

In the above relations, $R'_m$, $R'_a$, $R'_b$ are shown in figure 1.5.7, 1.5.8, and 1.5.9.

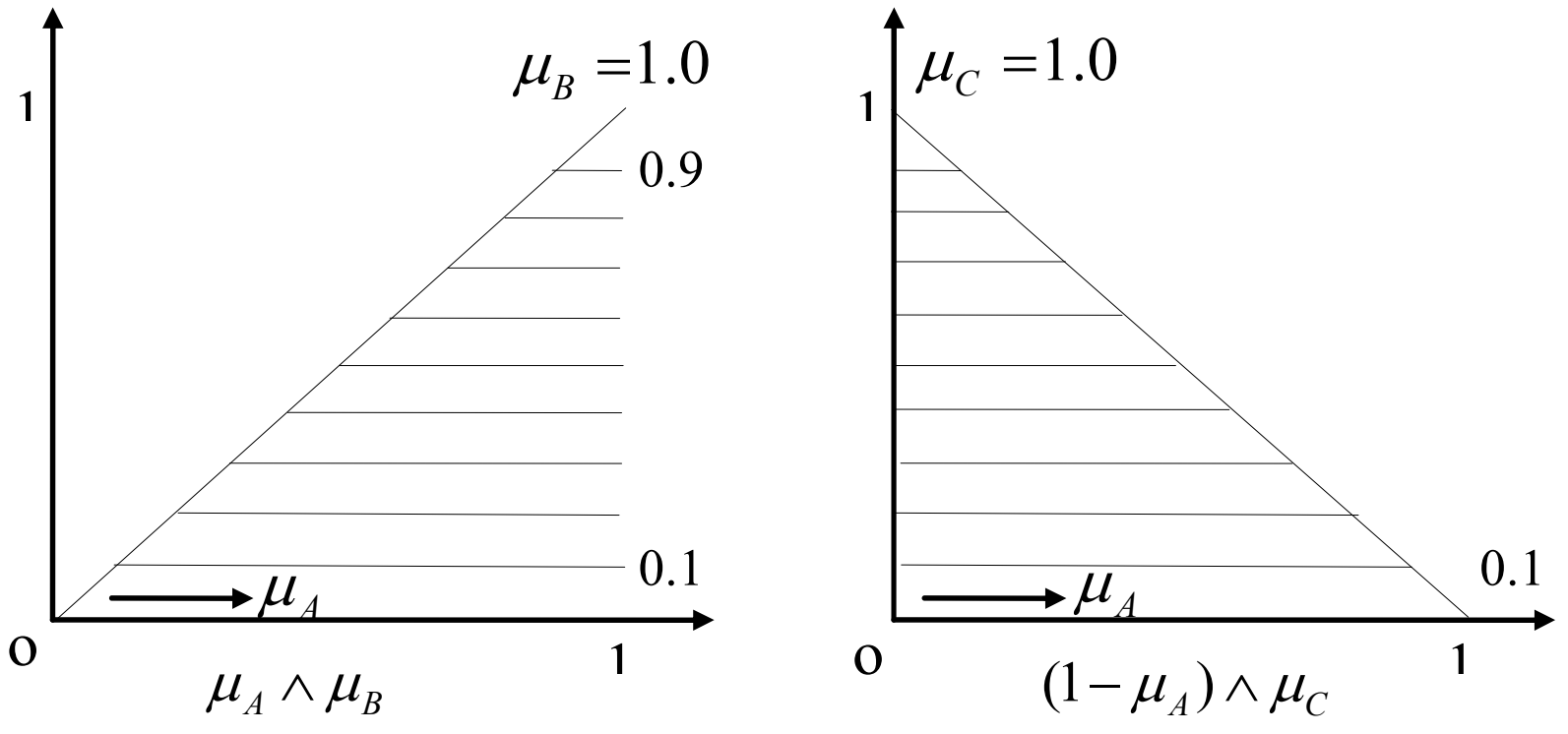

Figure 1.5.7 Fuzzy relation $R'_m$

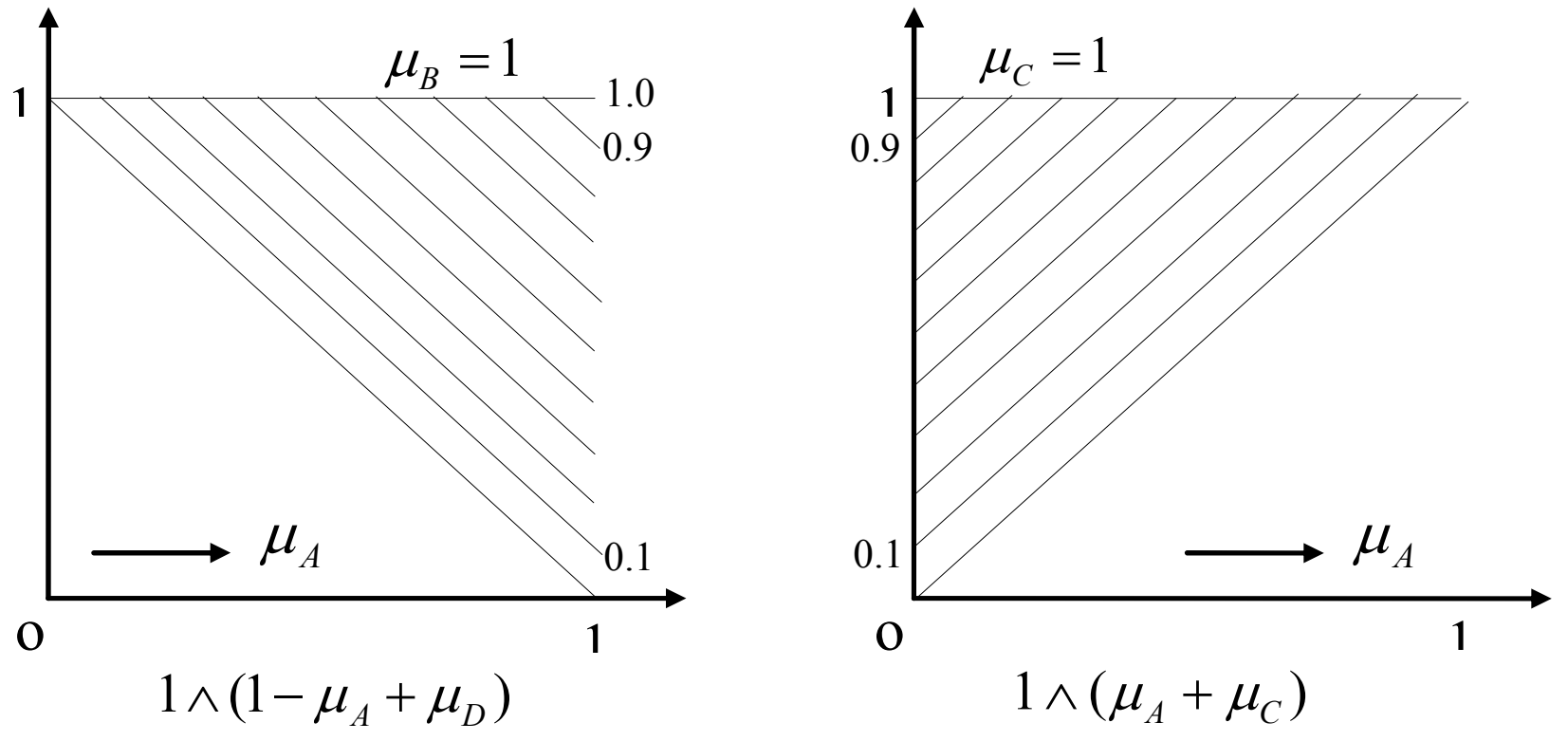

Figure 1.5.8 Fuzzy relation $R'_a$

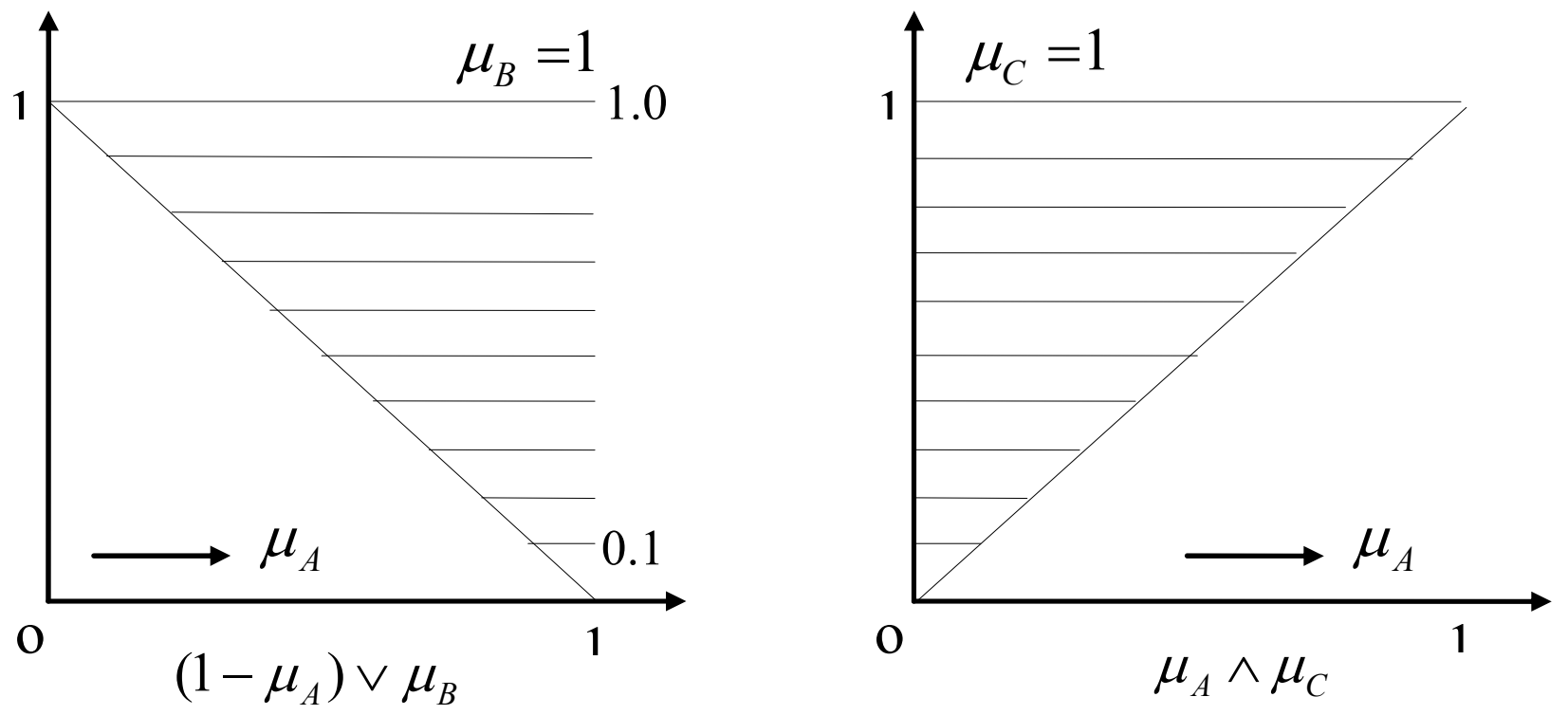

Figure 1.5.9 Fuzzy relation $R'_b$

(Example 1.5.2) $A'$ may be $A$, very $A(=A^2)$, more or less $A(=A^{0.5})$, not $A(=\neg A)$, not Very $A(=\neg A^2)$, not more or less $A(=\neg A^{0.5})$. For the convenience, we describe the case applying $A' = \text{very}A$ to $R'_m$.

In general, fuzzy relation $R$ is $R = R_1 \cup R_2$ in composition of fuzzy relation, then the following is satisfied.

$$A \circ (R_1 \cup R_2) = (A \circ R_1) \cup (A \circ R_2) \qquad (1.5.54)$$

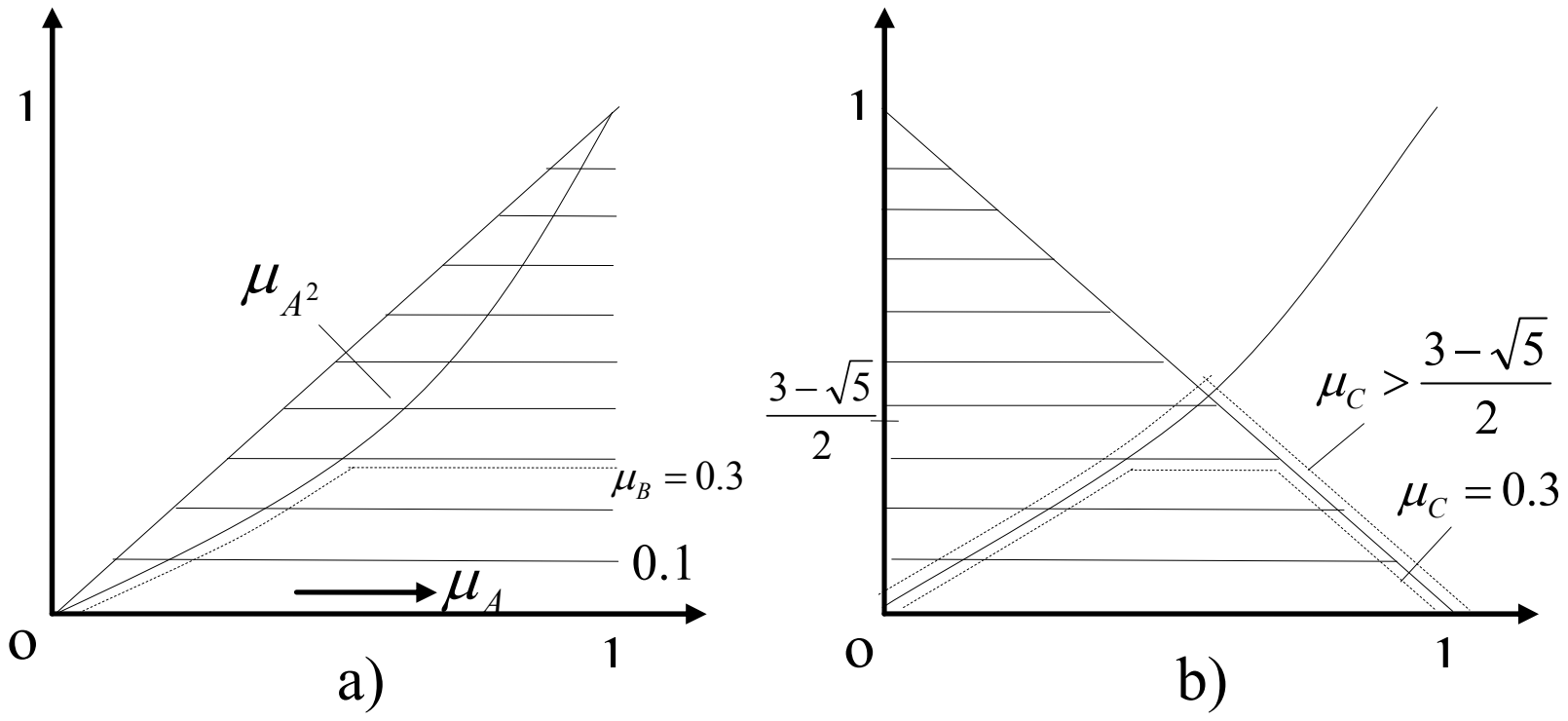

Figure 1.5.10 reasoning result by Fuzzy relation $R'_m$ in case $A^1 = A^2$

a) $\mu_{A^2 \circ (A \times B)}$, b) $\mu_{A^2 \circ (\neg A \times C)}$

Therefore, $D_m$ can be written as

$$D_m = A' \circ [(A \times B) \cup (\neg A \times C)] = [A' \circ (A \times B)] \cup [A' \circ (\neg A \times C)]$$

So, in case $A' = \text{very}A$, membership function of $A^2 \circ (A \times B)$ is

$$\mu_{A^2 \circ (A \times B)}(v) = \vee_u \{\mu_{A^2}(u) \wedge [\mu_A(u) \wedge \mu_B(v)]\}.$$

From this, if $\mu_B = 0.3$, then $\mu_{A^2} \wedge [\mu_A \wedge \mu_B]$ corresponds the dotted line in figure 1.5.10 a) and the maximum value is 0.3.

In general, the following is satisfied.

$$\mu_{A^2\circ(A\times B)} = \mu_B$$

Then, the membership function of $A^2 \circ (\neg A \times C)$ is

$$\mu_{A^2\circ(\neg A\times C)} = \vee_u \{\mu_{A^2} \wedge ((1-\mu_A)\wedge \mu_C)\}.$$

For example, if $\mu_C = 0.3\left(\le \frac{3-\sqrt{5}}{2}\right)$, then $\mu_{A^2} \wedge ((1-\mu_A)\wedge\mu_C)$ corresponds to the dotted line in the figure 1.5.10 b). From this, each maximum value is $\frac{3-\sqrt{5}}{2}$.

Therefore, in general

$$\mu_{A^2\circ(\neg A\times C)} = \begin{cases} \mu_C, \mu_C \le \frac{3-\sqrt{5}}{2} (\approx 0.382) \\ \frac{3-\sqrt{5}}{2}, \mu_C \ge \frac{3-\sqrt{5}}{2} \end{cases}$$

That is,

$$\mu_{A^2\circ(\neg A\times C)} = \frac{3-\sqrt{5}}{2} \wedge \mu_C$$

Then, the membership function of $D_m$ is as follows.

$$\mu_{D_m} = \mu_{A^2\circ(A\times B)\bigcup(\neg A\times C)} = \mu_B \vee \left(\frac{3-\sqrt{5}}{2} \wedge \mu_C\right)$$

Next, consider the case of $A^1 = \text{very}A$ and $R_a'$.

By $D_a = A^1 \circ [(\neg A\times V \oplus U\times B)\bigcap(A\times V\oplus U\times C)]$, the membership function of $D_a$ is as follows.

$$\mu_{D_a} = \underset{u}{\vee}\{\mu_{A^2} \wedge [1 \wedge (1-\mu_A + \mu_B) \wedge (\mu_A + \mu_C)]\}$$

However, $\mu_{A^2} \le \mu_A + \mu_c$, so the given equation becomes as follows.

$$\mu_{D_a} = \underset{u}{\vee}\{\mu_{A^2} \wedge [1 \wedge (1-\mu_A + \mu_B)]\}$$

This corresponds to the following equation and $(\neg A \times V \oplus U \times B)$ is the one converting the fuzzy conditional sentence 《If $x$ is $A$ then $y$ is $B$》 to the fuzzy relation.

$$A^2 \circ (\neg A \times V \oplus U \times B)$$

That is,

$$\mu_{D_a} = \frac{3 + 2\mu_B - \sqrt{5 + 4\mu_B}}{2}$$

(Example 1.5.3) Consider the case of $R_b'$.

Membership function of $R_b'$ is as follows by the definition and using $\mu_B$, $\mu_C$ as parameters, we can get as the minimum value I these two figures in figure 1.5.9.

$$((1-\mu_A) \vee \mu_B) \wedge (\mu_A \vee \mu_C)$$

For example, in case $\mu_C = 0.5,\ \mu_B = 0.2$, it corresponds to ① in figure 1.5.11 and in case $\mu_B = 0.7$, it corresponds to ②.

Therefore, from equation $D_b$, by $A^2$ and $R_b'$ the value of $\mu_{D_b}$ is the maximum height of dotted line, $\frac{3-\sqrt{5}}{2}$ in case $\mu_c = 0.5,\ \mu_B = 0.2$. $\frac{3-\sqrt{5}}{2}$ is the height of the crossing point of $\mu_{A^2}$ and $1-\mu_A$. In case $\mu_B = 0.7$, the maximum value is 0.7.

In case that $\mu_C$ changes, it is the same, therefore final reasoning result $\mu_{D_b} = \mu_B \vee \frac{3-\sqrt{5}}{2}$ is obtained.

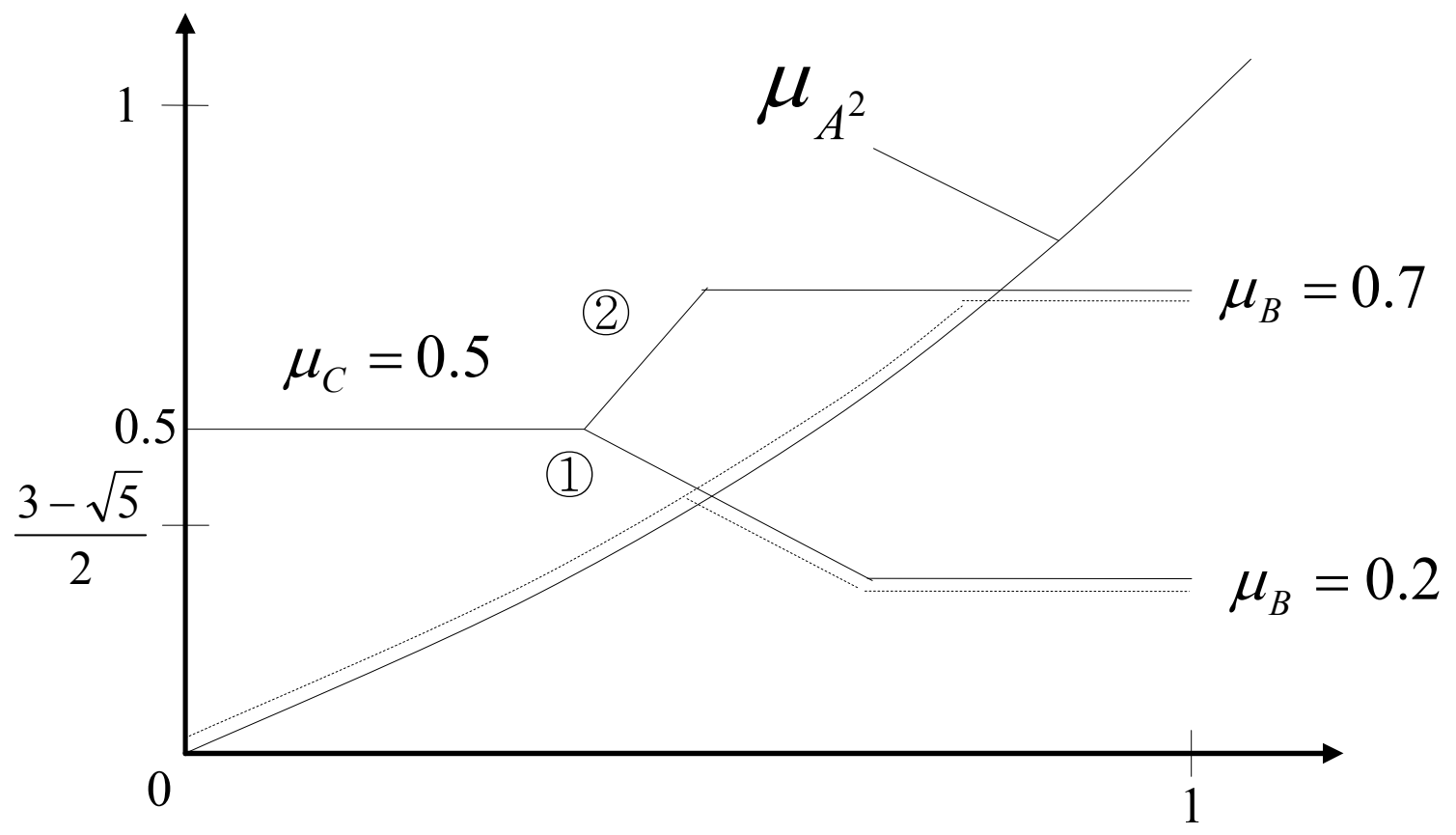

Figure 1.5.11 Reasoning result by Fuzzy relation $R'_b$ in case $A' = A^2$

In the same way, we can consider the cases that $A'$ is *A*, more or less *A*, not *A*, not Very *A*, not more or less *A*.

Applying the above cases to each method, the result is shown in table 1.5.2.

Table 1.5.2 IF-THEN-ELSE type fuzzy reasoning results

| | $A$ | very $A$ | more or less $A$ |
|---|---|---|---|
| $R'_m$ | $\mu_B \vee (0.5 \wedge \mu_C)$ | $\mu_B \vee \left( \frac{3-\sqrt{5}}{2} \wedge \mu_C \right)$ | $\mu_B \vee \left( \frac{\sqrt{5}-1}{2} \wedge \mu_C \right)$ |
| $R'_b$ | $\mu_B \vee 0.5$ | $\mu_B \vee \frac{3-\sqrt{5}}{2}$ | $\begin{cases} \mu_B; & \mu_B \geq (\sqrt{5}-1)/2 \\ & (\mu_B \vee \mu_C \vee 0.5) \\ \frac{\sqrt{5}-1}{2}; & \mu_B < (\sqrt{5}-1)/2 \end{cases}$ |
| $R'_a$ | $\frac{1+\mu_B}{2}$ | $\frac{3+2\mu_B-\sqrt{5+4\mu_B}}{2}$ | $\frac{\sqrt{5+4\mu_B}-1}{2} \wedge \frac{1+\mu_B+\mu_C}{2}$ |
| $R'_{gg}$ | $\mu_B$ | $\mu_B$ | $\begin{cases} \mu_B^{0.5}; & \mu_B+\mu_C \geq 1 \\ \mu_B \vee (\mu_B^{0.5} \wedge \mu_C); & \mu_B+\mu_C < 1 \end{cases}$ |
| $R'_{gs}$ | $\mu_B$ | $\mu_B$ | $\begin{cases}\mu_B^{0.5}; \mu_B+\mu_C \geq 1 \\ \mu_B; \mu_B+\mu_C < 1\end{cases}$ |
| $R'_{sg}$ | $\begin{cases}\mu_B; \mu_B+\mu_C \geq 1 \\ \mu_B \wedge \mu_C; \mu_B+\mu_C <\end{cases}$ | $\begin{cases}\mu_B^2; \mu_B+\mu_C \geq 1 \\ \mu_B^2+\mu_C; \mu_B+\mu_C < 1\end{cases}$ | $\begin{cases}\mu_B^{0.5}; \quad \mu_B+\mu_C \geq 1 \\ \mu_B^{0.5} \wedge \mu_C; \mu_B+\mu_C < 1\end{cases}$ |
| $R'_{ss}$ | $\begin{cases}\mu_B; \mu_B+\mu_C \geq 1 \\ 0; \mu_B+\mu_C < 1\end{cases}$ | $\begin{cases}\mu_B^2; \quad \mu_B+\mu_C \geq 1 \\ 0 \quad ; \quad \mu_B+\mu_C < 1\end{cases}$ | $\begin{cases}\mu_B^{0.5}; \quad \mu_B+\mu_C \geq 1 \\ 0 \quad ; \quad \mu_B+\mu_C < 1\end{cases}$ |
| $R'_n$ | $\mu_C \wedge (0.5 \wedge \mu_B)$ | $\mu_c \vee \left(\frac{3-\sqrt{5}}{2} \wedge \mu_B\right)$ | $\mu_C \vee \left(\frac{\sqrt{5}-1}{2} \wedge \mu_B\right)$ |
| $R'_b$ | $\mu_C \vee 0.5$ | $\mu_C \vee \frac{3-\sqrt{5}}{2}$ | $\begin{cases}\mu_C; \quad \mu_C \geq (\sqrt{5}-1)/2 \\ (\mu_C \vee \mu_B \vee 0.5) \\ \wedge \frac{\sqrt{5}-1}{2}; \ \mu_C < (\sqrt{5}-1)/2\end{cases}$ |
| $R'_a$ | $\frac{1+\mu_C}{2}$ | $\frac{3-\sqrt{5-4\mu_C}}{2}$ | $\frac{2\mu_C-1+\sqrt{5-4\mu_C}}{2} \wedge \frac{1+\mu_B+\mu_C}{2}$ |
| $R'_{gg}$ | $\mu_C$ | $\mu_C$ | $\begin{cases}1-(1-\mu_C)^2; \quad \mu_B+\mu_C \geq 1 \\ [1-(1-\mu_C)^2 \wedge \mu_B] \vee \mu_C; \ \mu_B+\mu_C < 1\end{cases}$ |
| $R'_{gs}$ | $\begin{cases}\mu_C; \mu_B+\mu_C \geq 1 \\ \mu_B; \mu_B+\mu_C < 1\end{cases}$ | $\begin{cases}1-(1-\mu_C); \mu_B+\mu_C \geq 1 \\ [1-(1-\mu_C)^{0.5}] \wedge \mu_B; \mu_B + \\ +\mu_C < 1\end{cases}$ | $\begin{cases}1-(1-\mu_C)^2; \quad \mu_B+\mu_C \geq 1 \\ [1-(1-\mu_C)] \wedge \mu_B; \ \mu_B+\mu_C < 1\end{cases}$ |
| $R'_{sg}$ | $\mu_c$ | $\mu_c$ | $\begin{cases}1-(1-\mu_c)^2; \quad \mu_B+\mu_c \geq 1 \\ \mu_c; \quad \mu_B+\mu_c < 1\end{cases}$ |
| $R'_{ss}$ | $\begin{cases}\mu_c; \mu_B+\mu_c \geq 1 \\ 0; \mu_B+\mu_c < 1\end{cases}$ | $\begin{cases}1-(1-\mu_c)^{0.5}; \mu_B + \\ +\mu_c \geq 1 \\ 0; \mu_B+\mu_c < 1\end{cases}$ | $\begin{cases}1-(1-\mu_c)^2; \ \mu_B+\mu_c \geq 1 \\ 0; \quad \mu_B+\mu_c < 1\end{cases}$ |

## 1.5.5 IF-THEN type multi fuzzy reasoning

Multi fuzzy reasoning is much used in fuzzy control and fuzzy expert system.

The type of multi fuzzy reasoning is as follows.

$$
\begin{array}{l}
1^{\circ}\ A_1 \text{ and } B_1 \to C_1 \\
2^{\circ}\ A_n \text{ and } B_2 \to C_2 \\
\quad \cdots \\
\underline{n^{\circ}\ A_n \text{ and } B_n \to C_n}
\end{array} \tag{1.5.55}
$$

Conclusion $C'$

Here, $A_i$, $A'$ are the fuzzy sets defined in $U$, $B_i$, $B'$ are in $V$ and $C_i$, $C'$ are in $W$.

If for $[A_i \text{ and } B_i]$, (1.5.56) is satisfied, then $[A_i \text{ and } B_i \to C_i]$ is expresses as (1.5.57)

$$\mu_{A_i \text{ and } B_i}(u,\ v) = \mu_{A_i}(u) \wedge \mu_{B_i}(v) \tag{1.5.56}$$

$$[\mu_{A_i}(u) \wedge \mu_{B_i}(v)] \to \mu_{C_i}(w) \tag{1.5.57}$$

For example, if $a \to b = a \wedge b$, then the above equation is as follows.

$$[\mu_{A_i}(u) \wedge \mu_{B_i}(v)] \wedge \mu_{C_i}(w)$$

Therefore,

$$
\begin{aligned}
C' &= (A' \text{ and } B') \circ [(A_1 \text{ and } B_1 \to C_1) \cup \cdots \cup (A_n \text{ and } B_n \to C_n)] = \\
&= [(A' \text{and} B') \circ (A_1 \text{and} B_1 \to C_1)] \cup \cdots \cup (A' \text{and} B') \circ (A_n \text{and} B_n \to C_n)] = \\
&= C_1' \cup \cdots \cup C_n'
\end{aligned} \tag{1.5.58}
$$

where $C_i' (i = 1,\ 2,\ \cdots,\ n)$ is

$$
\begin{aligned}
C_i' &= (A' \text{and} B') \circ (A_i \text{and} B_i \to C_i) = \\
&= [A_o'(A_i \to C_i)] \cap [B_o'(B_i \to C_i)]
\end{aligned} \tag{1.5.59}
$$

and its membership function is as follows.

$$\mu_{C_i'}(w) = \bigvee_u \{\mu_{A'}(u) \wedge \mu_{A_i}(u)\} \wedge \bigvee_v \{\mu_{B'}(v) \wedge$$

$$\wedge \mu_{B_i}(v)\} \wedge \mu_{C_i}(w) = a_i \wedge b_i \wedge \mu_{C_i}(w) \qquad (1.5.60)$$

This is shown in figure 1.5.12.

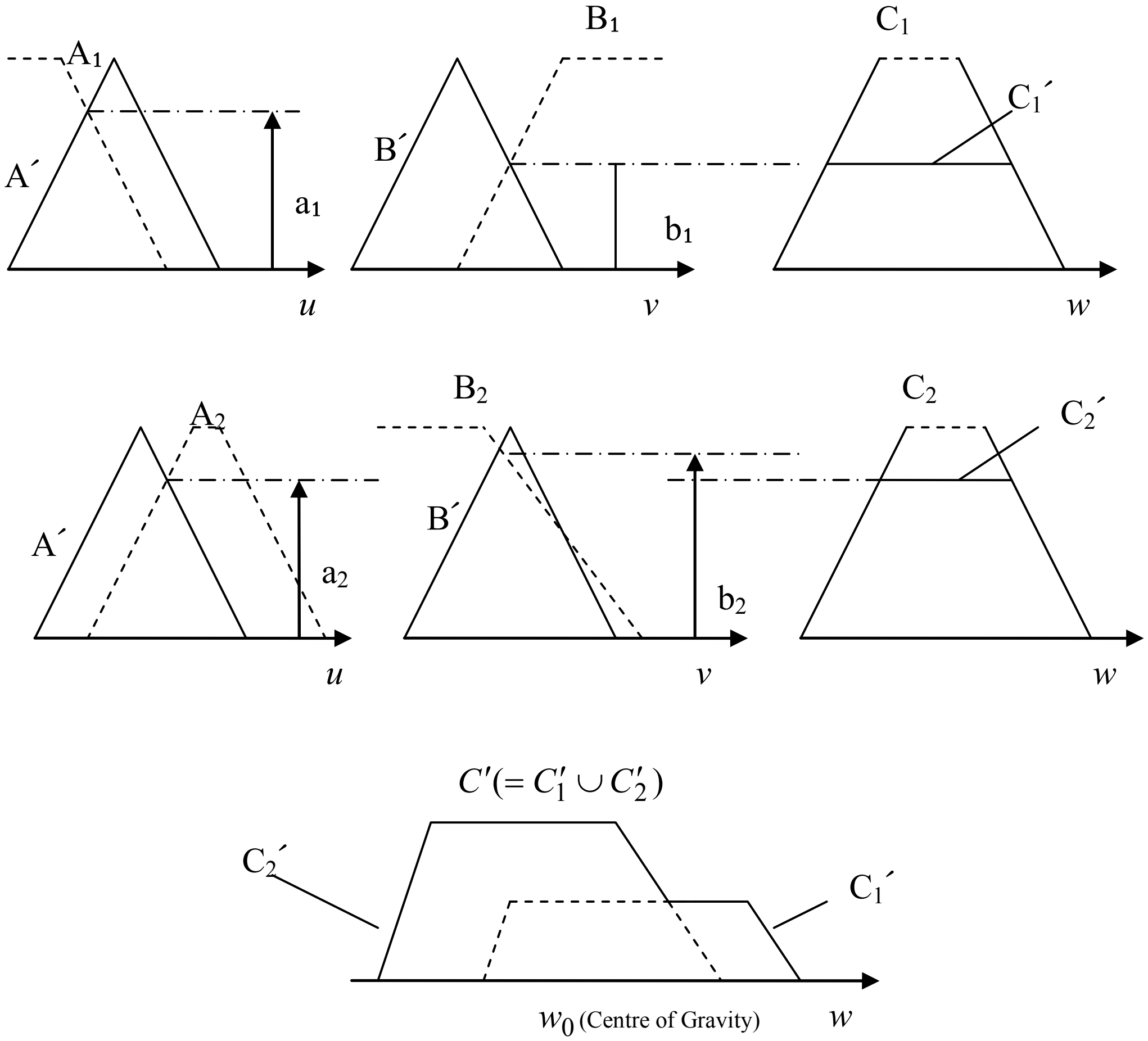

Figure 1.5.12 Multi fuzzy reasoning

And in case $A' = u_0, B' = v_0$, conclusion $C'$ is as (1.5.61) and $C_i'$ is as (1.5.62).

$$C' = C_1' \cup C_2' \cup \cdots \cup C_n' \qquad (1.5.61)$$

$$\mu_{C_i'}(w) = \mu_{A_i}(u_0) \wedge \mu_{B_i}(v_0) \wedge \mu_{C_i}(w) \qquad (1.5.62)$$

This is why $A'$ and $B'$ are observation values in fuzzy control.

## 1.5.6 Defuzzification method of fuzzy reasoning results

Fuzzy set obtained as the fuzzy reasoning results must be converted to the finite value (centre of gravity) to apply it to the practice.

The methods obtaining the centre of gravity $w_0$ are as follows.

$1^\circ$ Centre method

Select the centre of fuzzy set $C'$ as the centre of weight $w_0$. That is,

$$w_0 = \frac{\sum_i w_i \mu_{C'}(w_i)}{\sum_i \mu_{C'}(w_i)} \tag{1.5.63}$$

2° Maximum average method

Select the average value of the point with maximum class of fuzzy set $C'$ as the centre of gravity.

$$w_0 = \sum_{i=1}^{m} w_i / m \tag{1.5.64}$$

Here, $w_i'$ is the point having the maximum class and *m* is the number of it.

3° maximum middle point method

It is the special case of method 2°, and the middle point of minimum element $w'$ and maximum element $w''$ among the elements having the highest class as the centre of gravity.

$$w_0 = (w' + w'')/2 \tag{1.5.65}$$

4° Centroid method

Select the point dividing the area of fuzzy set $C'$ into two equal parts as the centre of gravity.

The examples for the above 4 methods are as follows.

(Example 1.5.4) Applying the Defuzzification method to the fuzzy set $C'$, the result is as follows.

$$w_0 = \frac{-2\times 0.4 + 1\times 0.8 + 0\times 0.6 + 1\times 0.8 + 2\times 0.8 + 3\times 0.2}{0.4+0.8+0.6+0.8+0.8+0.2} = 0.39$$

Maximum average method

$$w_0 = (-1+1+2)/3 = 0.67$$

Maximum middle point method

$w_0 = (1+2)/2 = 1.5$

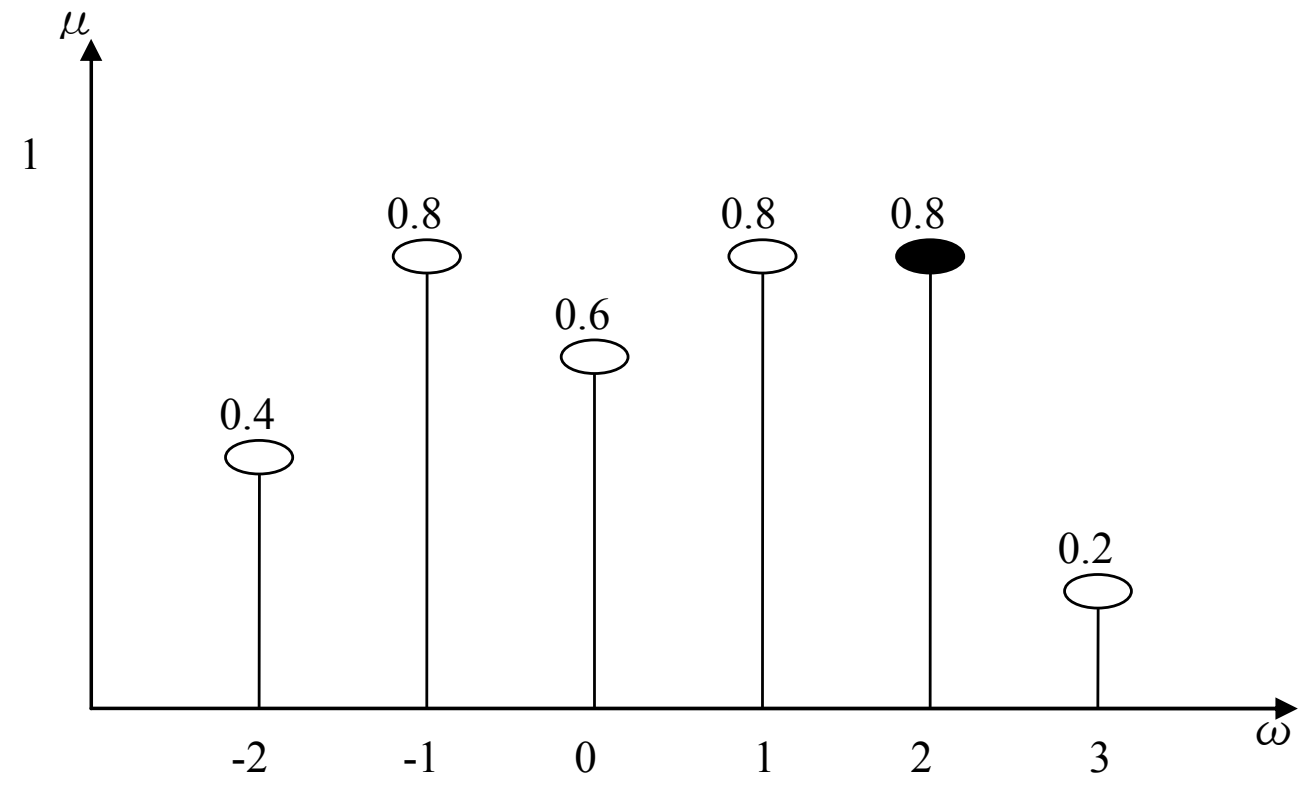

Figure 1.5.13 Centre method of fuzzy set $c''$

Centroid method

$$\omega_0 = 0.5\begin{pmatrix} left\ hand = 0.4+0.8+0.6 \\ right\ hand = 0.8+0.8+0.2 \end{pmatrix}$$

The following centre of gravity method (Defuzzification method) is not to get the center of gravity from the mixed fuzzy set $C'$, but is the method to get it, by using the characteristics of fuzzy sets $C_1', C_2', \cdots, C_n'$ (height, area) reasoned from each reasoning rule.

In figure 1.5.14, referring the height of the reasoned fuzzy set $C_i'$ to $h_i$,, and the area to $S_i$, and the centre of gravity to $w_i$, the following centre of gravity is considered, where the height $h_i$ is as follows.

$$h_i = \mu_{A_i}(x_0) \wedge \mu_{B_i}(y_0)$$

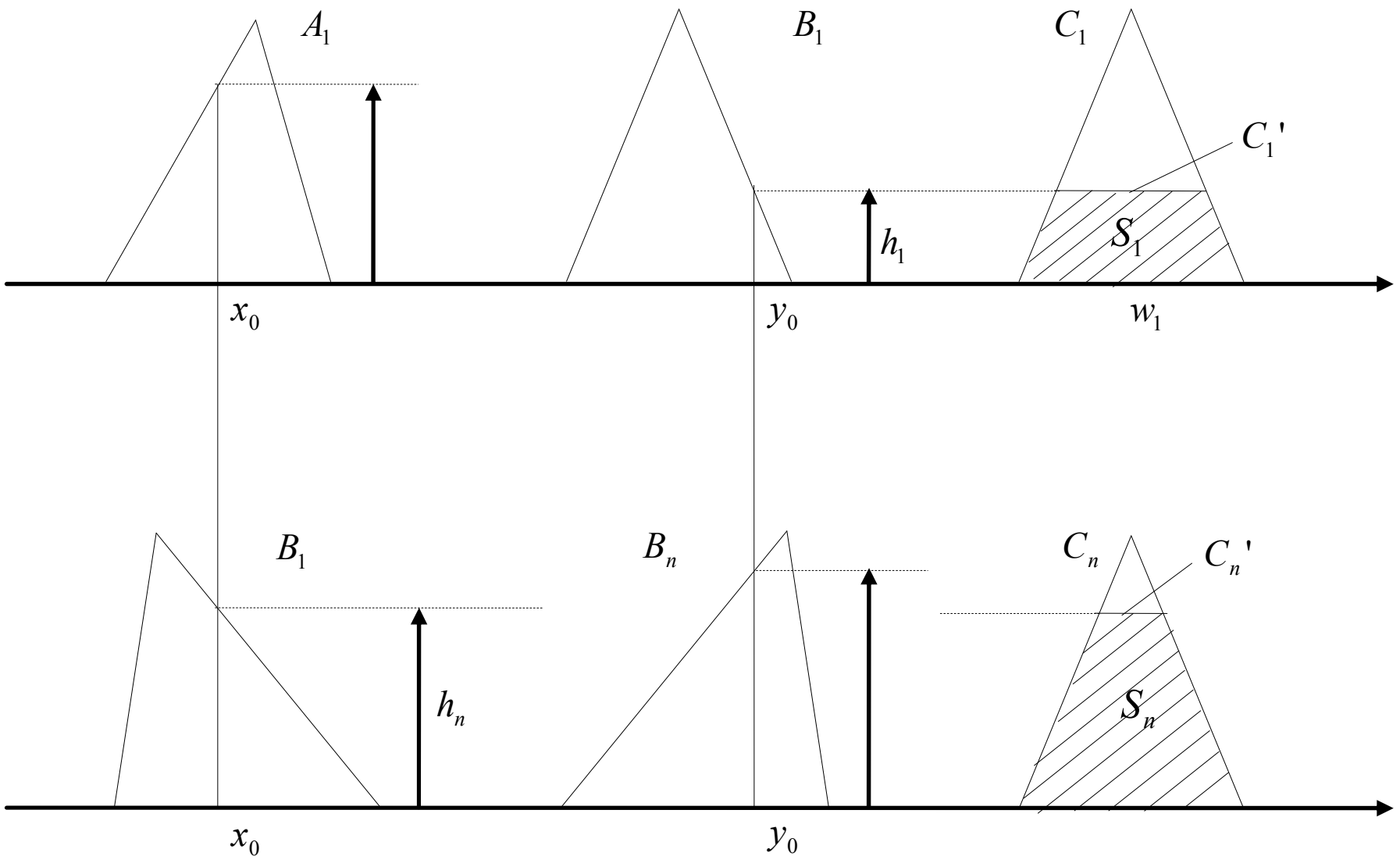

Figure 1.5.14 Height $h_i$ and area $S_i$ of reasoning result $C_i'$

5° Height method

Average the representation point $z_i$ of $C_i$ into the height $h_i$ of fuzzy set $C_i'$.

$$\omega_0 = \frac{w_1 h_1 + w_2 h_2 + \cdots + w_n h_n}{h_1 + h_2 + \cdots + h_n} \qquad (1.5.66)$$

6° Maximum height method

$$w_0 = w_i (h_i \text{ is the maximum height}) \qquad (1.5.67)$$

7° Area method

Average the representation point $w_i$ of $C_i$ into the area $S_i$ of fuzzy set $C_i'$.

$$w_0 = \frac{z_1 s_1 + z_2 s_2 + \cdots + z_n s_n}{s_1 + s_2 + \cdots + s_n} \qquad (1.5.68)$$

8° Greatest area method

Select the centre of gravity of $C_i$ $w_i$ for the maximum area $S_i$ as the centre of gravity of the reasoning result.

### 1.5.7 Fuzzy reasoning by various compositional rules

Up to now, since fuzzy reasoning 《If *x* is *A* then *y* is *B*》 expresses a certain fuzzy relation between *A* and *B*, convert the fuzzy conditional sentence into fuzzy relation *R*, max-min composition of $A'$ and *R*, and then get the conclusion $B'$.

Here, we consider the method of obtaining the result $B'$, by composing the *R* and $A'$ on the basis of max－⊙ and max－$\wedge$ instead of composition max－min.

① Compositional rule □ of max－⊙

$$A□R \Leftrightarrow \mu_{A?R}(v) = \vee_u \{\mu_A(u) \odot \mu_R(u,v)\} \tag{1.5.69}$$

where operation ⊙ is defined as follows for $x,\ y \in [0,\ 1]$.

$$x \odot y = 0 \vee (x + y - 1) \tag{1.5.70}$$

When $A'$ is given for reasoning rule $R_m$, $B'_m$ is expressed as follows by compositional rule □.

$$B'_m = A' □ R_m = A' □ [(A \times B) \cup (\neg A \times V)], \tag{1.5.71}$$

$$\mu_{B'_m}(v) = \vee_u \{\mu_A \cdot (u) \odot [(\mu_A(u) \wedge \mu_B(v)) \vee (1 - \mu_A(u))]\} \tag{1.5.72}$$

And given $B'$, $A'_m$ is expressed as follows.

$$A'_m = R_m □ B', \tag{1.5.73}$$

$$\mu_{A'_m}(u) = \vee_v \{[(\mu_A(u) \wedge \mu_B(v)) \vee (1 - \mu_A(u))] \odot \mu_{B'}(v) \tag{1.5.74}$$

② Compositional rule of $\max - \wedge$ ▲

$$\mathrm{A}\blacktriangle\mathrm{R} \leftrightarrow \mu_{A\,\blacktriangle\,R}(v) = \bigvee_{u}\{\mu_A(u) \mathbin{\hat{\wedge}} \mu_R(u,\ v)\} \tag{1.5.75}$$

Operation $\hat{\wedge}$ is defined as follows for $x,\ y \in [0,\ 1]$.

$$x \mathbin{\hat{\wedge}} y = \begin{cases} x;\ y = 1 \\ y;\ x = 1 \\ 0;\ 0 < x,\ y < 1 \\ x \wedge y;\ x \dot{+} y = 1 \\ 0;\ 0 \le x \dot{+} y < 1 \end{cases} \tag{1.5.76}$$

where $x \dot{+} y = x + y - x \cdot y$

For $R_m$ and $A'$,

$$B'_m = A' \blacktriangle R_m \tag{1.5.77}$$

$$\mu_{B'_m}(v) = \bigvee_{u}\{\mu_{A'}(u) \mathbin{\hat{\wedge}} [(\mu_A(u) \wedge \mu_B(v)) \vee (1 - \mu_A(u))]\} \tag{1.5.78}$$

For $R_m$ and $B'$

$$A'_m = R_m \blacktriangle B' \tag{1.5.79}$$

$$\mu_{A'_m}(u) = \bigvee_{v}\{[(\mu_A(u) \wedge \mu_B(v)) \vee (1 - \mu_A(u))] \mathbin{\hat{\wedge}} \mu_{B'}(v)\} \tag{1.5.80}$$

As we discussed in the above fuzzy reasoning, we can consider the reasoning result in case that $A'$ is

$$\begin{aligned} &A' = A \\ &A' = \mathrm{very}A = A^2 \\ &A' = \mathrm{moreorless}A = A^{0.5} \\ &A' = \mathrm{not}A = \neg A \end{aligned}$$

and $B'$ is

$$\begin{aligned} &B' = \mathrm{not}B = \neg B \\ &B' = \mathrm{notvery}B = B^2 \\ &B' = \mathrm{notmoreorless}A = \neg B^{0.5} \\ &B' = B \end{aligned}$$

(Example 1.5.5) Given $A^1 = A$, consider the fuzzy reasoning process by new compositional rule in case of fuzzy relation $R_m$.

Obtain the reasoning result $B'_m$ in case $A' = A$ by □ and ▲.

Here, assuming that $\mu_A$ is the function in $[0, 1]$, then it can be rewritten as

$$b'_m = \vee_x \{x \odot [(x \wedge b) \vee (1-x)]\}$$

where

$$x = \mu_A(u)$$
$$b = \mu_B(v)$$
$$b'_m = \mu_{B'_m}(v)$$

Using $f(x) = x \odot [(x \wedge b) \vee (1-x)]$, $f(x)$ becomes the following by the definition of $\odot$.

$f(x) = x \odot [(x \wedge b) \vee (1-x)] = 0 \vee \{x + [(x \wedge b) \vee (1-x)] - 1\} =$

$= 0 \vee \{[x - 1 + (x \wedge b)] \vee (x - 1 + 1 - x)\} = 0 \vee \{[(x - 1 + x) \wedge (x - 1 + b)] \vee 0\} =$

$= [0 \vee (2x-1)] \wedge [0 \vee (x - 1 + b)]$

Figure 1.5.15 was drawn with $0 \vee (2x-1)$ and $0 \vee (x-1+b)$ as parameters.

In figure, if $b = 0.2$, then $f(x)$ corresponds to the dotted line and the maximum value is 0.2.

Therefore, if $b = 0.2$, then $b'_m = \vee_x f(x) = 0.2$.

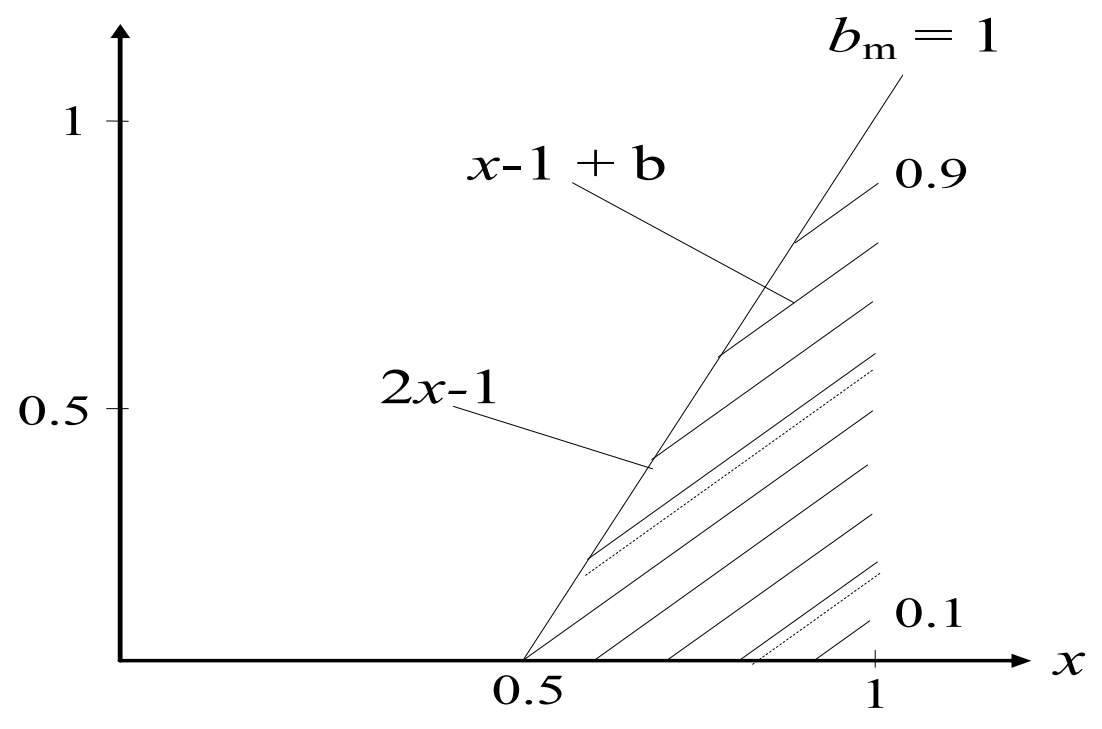

Figure 1.5.15. $0 \vee (2x-1)$ and $0 \vee (x-1+b)$

In the same way, if $b = 0.6$, then $b'_m = 0.6$. Therefore, if $x' = x$, then $b'_m = b$.

That is,

$$A \,\square\, R_m = B$$

and $R_m$ satisfies the modus ponens for $\max-\odot$ composition □.

But, $R_m$ doesn't satisfy modus ponens for $\max-\min$ composition.

Next, consider whether compositional rule □ and ▲ satisfy the syllogism.

Referring $P_1$, $P_2$, $P_3$ to fuzzy conditional sentences, syllogism is as follows.

$$P_1\text{: If } x \text{ is } A \text{ then } y \text{ is } B$$
$$P_2\text{: if } y \text{ is } B \text{ then } z \text{ is } C$$
$$P_3\text{: if } x \text{ is } A \text{ then } z \text{ is } C$$

$R(A, B)$, $R(B, C)$, $R(A, C)$ are the fuzzy relations in $U \times V$, $V \times W$, $U \times W$, which are obtained from fuzzy conditional sentences $P_1$, $P_2$, $P_3$.

If the following is satisfied for compositional rule (1.5.81), $*\ (\in \{\square, \blacktriangle\})$, then it is said that syllogism is held.

$$R(A,B) * R(B,C) = R(A,C) \qquad (1.5.81)$$

For example, in case $* = \square$, the membership of $R(A,B) \,\square\, R(B,C)$ becomes as follows.

$$\mu_{R(A,B)\square R(B,C)}(u,w)=\underset{v}{\vee}\{\mu_{R(A,B)}(u,v)\odot\mu_{R(B,C)}(v,w)\}$$

(Example 1.5.6) For fuzzy relation $R_b$, consider whether syllogism is satisfied applying composition □ according to (1.5.81).

By $R_b$,

$$\mu_{R_b(A,B)}(u,\ v)=(1-\mu_A(u))\vee\mu_B(v)$$
$$\mu_{R_b(B,C)}(v,\ w)=(1-\mu_B(v))\vee\mu_C(w)$$

$$d=\mu_{R(A,B)\square R(B,C)}(u,w)$$
$$a=\mu_A(u),\ x=\mu_B(v),\ C=\mu_C(w)$$

and if $\mu_B$ is the function in $[0,\ 1]$, then

$$d=\underset{x}{\vee}\{[(1-a)\vee x]\odot[(1-x)\vee C]\}$$

$$f(x)=[(1-a)\vee x]\odot[(1-x)\vee C]$$

Therefore,

$$f(x)=0\vee\{[(1-a)\vee x]+[(1-x)\vee c]-1\}=0\vee(1-a-x)\vee(x+c-1)\vee(c-a)$$

1° In case $c-a\geq 0$

2° In case $c-a\leq 0$

Therefore, for composition □, $R_b$ satisfies the syllogism.